%% file: FSPP.tex
\documentclass[%
	pdftex,
	a4paper,
	oneside,
	bibtotocnumbered,
	idxtotoc,
	halfparskip,
	chapterprefix,
	headsepline,
	footsepline,
	12pt
]{scrbook}

\pdfoutput=1

\usepackage{makeidx}

\usepackage[latin1]{inputenc}
\usepackage[T1]{fontenc}

\usepackage{array}

\usepackage{graphicx}

\setkomafont{sectioning}{\normalfont\bfseries}
\setkomafont{captionlabel}{\normalfont\bfseries}
\setkomafont{pagehead}{\normalfont\itshape}
\setkomafont{descriptionlabel}{\normalfont\bfseries}

\setcapindent{1em}

\usepackage{amsmath}
\usepackage{amssymb}

\usepackage[scaled=.92]{helvet} 

\usepackage{rotating}

\usepackage{color}

\definecolor{LinkColor}{rgb}{0,0,0.5}

\usepackage{hyperref}

\usepackage[savemem]{listings}
	{\begin{list}{$\diamondsuit$}{}}%
	{\end{list}}

\newcommand{\qsr}{qualitative spatial reasoning}
\newcommand{\tpcc}{Ternary Point Configuration Calculus}
\newcommand{\doi}{Distance/orientation-interval propagation}
\newcommand{\csp}{Constraint satisfactory problems}
\newcommand{\AI}{Artificial Intelligence}
\newcommand{\sk}{spatial knowledge}
\newcommand{\si}{spatial information}
\newcommand{\FSPP}{Fine-grained Qualitative Spatial Reasoning about Point Positions}
\newcommand{\fs}{\textsc{Fspp}}

\begin{document}

\input{chapters/titlepage.tex}

\input{chapters/abstract}

\tableofcontents
\listoffigures

\input{chapters/introduction.tex}

\input{chapters/foundation.tex}

\input{chapters/stateOfArt.tex}

\input{chapters/FSPPapproach.tex}



\input{chapters/conclusion.tex}

\nocite{cosy:dylla:2004:TPCC_Complexity}
\nocite{Pavlidis}

\nocite{hernandez1997}
\nocite{hernandez1995}

\appendix
\input{appendix/TPCC.tex}

\input{appendix/DOI.tex}

\input{appendix/RCC.tex}

\bibliographystyle{bib/apalike}

\bibliography{bib/diploBib}


\end{document}

%% file: chapters/titlepage.tex
%
%
%
%
\begin{titlepage}
 
  \thispagestyle{empty}%
  \voffset=1in
  \topmargin=0pt
  \headheight=0pt
  \headsep=0pt

\setlength\paperheight{397mm}
\setlength\textheight{397mm}
\setlength\footskip{-100mm}
  \parskip=0pt
  \parindent=0pt

  \newpage

\begin{center}
		\includegraphics[width=3cm]{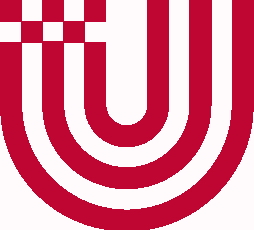}\\
{\large		University of Bremen\\
		Department of Computer Science}\\

\end{center}

	\begin{center}

	\vspace*{2cm}

		{\Large Diploma Thesis}\\\vspace{1cm}
		{\LARGE
	\textbf{Fine-grained Qualitative Spatial Reasoning about Point Positions}}\\
		\vspace{1.5cm}
		{\Large
		S\"{o}ren Schwertfeger\\
		August 2005\\\vspace{1.5cm}
		
		Supervisors:\\
		Prof. Christian Freksa, Ph.D.\\
	  Dr. Reinhard Moratz}\\
  
  		\includegraphics[width=14cm]{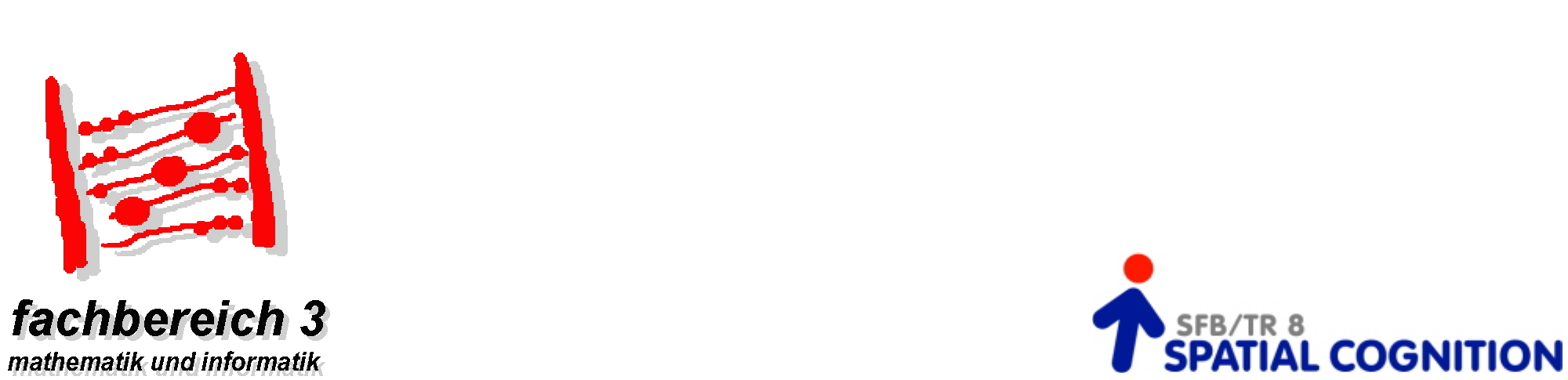}



	\end{center}
	\normalsize
	\vfill

\end{titlepage}

%% file: chapters/abstract.tex
%
%
%
%
\thispagestyle{empty}

\section*{}
\newpage
\thispagestyle{empty}

\textbf{\Large Abstract}\\\vspace{0.5cm}

The ability to persist in the spacial environment is, not only in the robotic context, an essential feature.  Positional knowledge is one of the most important aspects of space and a number of methods to represent these information have been developed in the in the research area of spatial cognition. The basic qualitative spatial representation and reasoning techniques are presented in this thesis and several calculi are briefly reviewed. Features and applications of qualitative calculi are summarized. A new calculus for representing and reasoning about qualitative spatial orientation and distances is being designed. It supports an arbitrary level of granularity over ternary relations of points. Ways of improving the complexity of the composition are shown and an implementation of the calculus demonstrates its capabilities. Existing qualitative spatial calculi of positional information are compared to the new approach and possibilities for future research are outlined.

\newpage
\thispagestyle{empty}

\section*{}
\newpage
\thispagestyle{empty}

\textbf{\Large Acknowledgments}\\\\
First of all, I would like to thank my advisor Dr. Reinhard Moratz for introducing me to the interesting field of qualitative spacial representation and reasoning, for the inspiring discussions and his support. Without him this thesis would not have been possible. Furthermore I want to express my gratitude to Prof. Christian Freksa for supervising this thesis.\\
I am thankful to all members of the spatial cognition group for the creative atmosphere that I enjoyed there. 
As the five years as a student at the University of Bremen pass by I would like to thank my friends and classmates for making these years a wonderful experience.\\ 
Finally I thank my family for their support, especially my daughter for brightening up my day and my wife for her constant backup.\\\vspace{1cm}


%% file: chapters/introduction.tex
%
%
%
%

\chapter{Introduction}
\label{sec:Introduction}

%
%
%
%
%
\section{Motivation}
Space is a basal part of the existence of all creatures. The ability to cope with the spatial environment is thus a primary survival skill of even the smallest insects.\\
All physical entities are located in space and their spatial properties, like location, orientation or shape, might change. Because humans are able to deal with spatial knowledge effectively we can deliberately and intelligently utilise this environment. We obtain and modify this knowledge by what we perceive from the external world and by our ability to derive new knowledge. The communication of spatial knowledge is an essential skill of human beings.\par

The elementary concept of space has thus been a central subject of study in Cognitive Science and \AI. To perceive space, to represent space, to reason about space and to communicate space is an important task in \AI~research. \AI~ is a highly diversified discipline with influences from different classical areas such as Computer Science, Logic, Engineering, Psychology and Philosophy.\par

The two primary methods to deal with space are the quantitative approach and the qualitative approach. The quantitative approach is the classical way which is used in many traditional disciplines like computational geometry, computer vision and robotics. It uses numerical values within a coordinate system to store the \si. The qualitative methods abstract from the exact numerical values and use a finite vocabulary to handle \si. Qualitative representation of \sk~ and reasoning about qualitative \sk~ is more "human-like" or "cognitively adequate" than classical approaches.\par

\AI~and in particular spatial reasoning have many applications, but an increasingly important one is robotics. Non-static robots most often solve spatial problems like maintaining information about their position, path-finding, way-planning, map-building or goal-finding. Communicating with humans about \sk~is another important aspect of robotics. Topological information are often useful, but for robotics the most important \si~ is the positional information. Qualitative approaches are very suited for dealing with \sk~ in robotics because they can handle imprecise sensor data and omit unnecessary details.\par

The goal of this thesis is to develop a new qualitative calculus for spatial reasoning about positions. Most often it is sufficient to reason about points and to abstract from the extend of the objects. Many existing qualitative spatial representations are too coarse for some applications, especially for robotics. Thus a calculus for more "Fine-grained Qualitative Spatial Reasoning about Point Positions" is being presented in this thesis. This thesis is written with regard to robotics but the calculus can be used for other applications that need to reason about positional information as well.\par

\section{Structure of this thesis}
This thesis will first introduce the reader to the foundations of qualitative spatial reasoning. State of the art calculi are presented in following chapter. After the new approach is developed, ways of improving it and it's implementation are explained. A summary followed by the discussion and the outlook is presented after a comparison of the new calculus with other approaches. 

The first section of Chapter \ref{sec:Foundations}, which introduces the terms and definitions required, is about spatial representations in general. It presents different important aspects of space like topology, orientation, distance and shape. Section \ref{sec:spatialReasoning} introduces to spacial reasoning and the unary and binary operations that are needed for it. General constraint based reasoning is explained as well as the special features of spatial reasoning. The next Section (\ref{sec:qualitative_spatial_calculi}) presents the qualitative representation, its properties and it gives reasons why to use it. Later in this section qualitative spatial representation and reasoning is introduced. The last section of Chapter \ref{sec:Foundations} (Section \ref{sec:applications}) is about the applications in which qualitative spatial reasoning is used.

Chapter {sec:StateOfTheArt} introduces state of the art calculi. The first calculus in Section \ref{sec:rcc} is about topological aspects of space. The Dipole approach over intrinsic orientation knowledge is presented in Section \ref{sec:dipole}. Section \ref{sec:tpcc} introduces the \tpcc~from which the calculus developed in this thesis derives. The \doi~is a quantitative approach which is used for the computation of the composition (Section \ref{sec:doi}). The Granular Point Position Calculus introduced in Section \ref{sec:gppc} is recently developed and similar to this thesis's approach.

The calculus of this diploma thesis is being developed in Chapter \ref{sec:FSPP approach}. The absolute distance representation is presented in Section \ref{sec:fsppDist} and the relative orientation representation is introduced in Section \ref{sec:fsppOri}. Section \ref{sec:fsppPos} combines those latter two to represent positional knowledge. The reasoning techniques for the new calculus are explained in Section \ref{sec:fsppReas} and the conceptual neighborhoods are defined. The algorithm for the composition as well as the unary operations are developed and improved in this section, for example by using Pavlidis contour tracing algorithm. Section \ref{sec:impl} shows code fragments of the implementation as well as the output of an example program while Section \ref{sec:comparison} compares some important state of the art calculi with the newly developed one.

Chapter \ref{sec:conclusion} provides a summary of the results achieved in this thesis as well as a discussion and an outlook for future research possibilities. 

The appendix contains in depth definitions for the \tpcc~(\ref{cap:TPCCdef}), for the \doi~(\ref{cap:DOIcompform}) and for the Region Connection Calculus (\ref{app:topology}) as well as the Bibliography.

%% file: chapters/foundation.tex
%
%
%
%


\chapter{Foundations}
\label{sec:Foundations}

%
%
%
%
%






%
%
%
%
%







%
%
%
%
%
\section{Spatial Representation}
\label{sec:rep1}
There are tasks within robotics that can be performed without the need of spatial representations. Behavior-based robotics does not use an internal model of the environment. The robot rather acts by directly using the input from the sensors and thus reacting to the changes in its environment. One very early example of behavior-based robotics is Walter's turtle. There, a light indicates the position of a recharge station. The turtle has a photo sensor which leads it to the recharge station if the batteries run low on power and makes it flee from the light once the batteries are full again, \cite{Walter}. Other great contributions to behavior-based robotics have been done by Braitenberg who developed the famous Braitenberg-vehicles \cite{Braitenberg} as well as by Ronald Arkin \cite{Arkin}.\par





More complex demands on robots require the use of \sk. In order to use this \sk~ and perform tasks like reasoning with this information it is necessary to have an adequate representation. There are two common ways to represent \sk: the qualitative and the quantitative (it is also known as numerical) approach which are closely introduced in section \ref{sec:qualitative_spatial_calculi}. \par

There are different aspects of space that are represented by different kinds of spatial relationships. The one dimensional Temporal logic will be presented briefly as it had great influence on \qsr. Topology representations are inherently qualitative whereas orientation and distance information can also be expressed quantitatively. The latter two combined form the positional information. Those aspects of space are often not independent from each other, especially if orientated objects with extension are represented.

\subsection{Temporal Logic}
\label{subsec:templogic}
James F. Allen introduced a temporal logic in 1983 based on intervals, \cite{Allen}. Without using the term "qualitative", Allen defined some conditions for his algorithm: It should allow imprecision (relative relations rather than absolute data are used) and uncertainty of information (constraints between two times may exist although the relationship between two times may be unknown). Furthermore the algorithm should vary in its grain of reasoning (years, days, milliseconds etc.) and it should support persistence (facilitate default reasoning).\par

Allen uses intervals of time because he observed that there don't seem to be any atoms in time. All possible events in time might be broken up into two or more individual events thus forming an interval of time. 

\begin{table}
	\centering
		\begin{tabular}{c|c|c|l}
			Relation	& Symbol & Inverse symbol & Example\\ \hline
			X equal Y & = & = & XXX\\
				& & & YYY\\\hline
		  X before Y& < & > & XXX~~~YYY\\\hline
			X meets Y & m & mi & XXXYYY\\\hline
			X overlaps Y & o & oi & XXX\\
			& & & ~~YYY\\\hline
			X during Y & d & di & ~~XXX\\
			& & & YYYYY\\\hline
			X starts Y & s & si & XXX\\
			& & & YYYYY\\\hline
			X finishes Y & f & fi & ~~XXX\\
			& & & YYYY\\
		\end{tabular}
	\caption{Thirteen Relationships}
	\label{tab:timerelations}
\end{table}

In table \ref{tab:timerelations} we can see the thirteen possible relationships between intervals that Allen identified. Convenience  allows the collapse of the three during relations (d, s, f) into one relationship called dur and the three containment relations (di, si, fi) into a relationship called con.\par

The time intervals constitute nodes of a network, while the arcs between those nodes are labeled with one or more of those thirteen relationships (allowing uncertainty for disjunction of relations). Allen presented an algorithm to calculate the transitive closure of such a network using a "transitivity table" (composition table). He also introduced reference intervals to group clusters of intervals and thus reduced the space requirements of the representation without greatly affecting the inferential power of his mechanism. \cite{Allen}\par

A generalization of Allen's approach was introduced in 1990 by Christian Freksa,\\\cite{FreksaTemporal}. Rather than reasoning about intervals of time he used semi-intervals. The start and end-points (which themselves can be seen as semi-intervals on higher levels of granularity) now define the intervals using the relations before (<), equal (=) and after (>). Thus all relations of intervals can be uniquely defined by using a maximum of two relations of beginnings and endings. This is possible because of two domain-inherent conditions: beginnings take place before endings and the relations (<, =, >) are transitive.\par

Freksa's approach uses the "Conceptual Neighborhood" of relations which allows for inferring neighbors of relations between objects about which neighborhoods of relations to some other object are known. It is more efficient in regards of inferencing while the full reasoning power is maintained, thus this calculus is cognitively plausible. We see, for example, that having less knowledge corresponds to a simpler representation rather than Allen's method, which adds more relation disjunctions, if less is known. This also allows for a drastic compaction of the inferencing knowledge base.\par

\cite{FreksaTemporal}\par

Efforts have been made to extend temporal logic into spatial dimensions such as using the temporal distinctions for the two axes of a Cartesian coordinate system \cite{Guesgen}. These approaches lack cognitive plausibility because humans don't decompose the world into two axes nor consider relations on each of them. 

It must be noted that there are (although both are one dimensional) important differences between time and a one-dimensional space. The most important is, that time always moves forward whereas space has no fixed direction. Therefore spatial representations differ greatly from temporal logic.

\subsection{Topology}
Topological distinctions between spatial entities are a fundamental aspect of spatial knowledge. Because those distinctions are inherently qualitative they are particularly interesting for \qsr. 

Topological representations usually describe relationships between spatial regions which are subsets of some topological space. Most approaches that formalize topological properties of spatial regions use a single primitive relation (the binary connectedness relation) to define many other relations. In topology, formal definitions like neighborhood, interior, exterior, boundary etc. can be used to define relations like 
disjoint, meet, overlap, contain etc. for two regions. \\

\begin{figure}[hb]
	\begin{captionabove}{Examples for topological relations}
		\label{fig:topo}
\begin{center}
		\includegraphics[width=7cm]{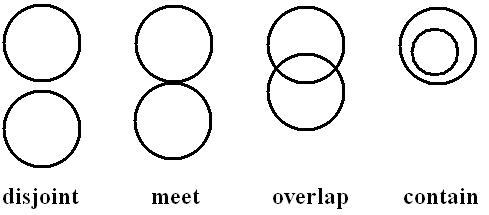}
\end{center}
	\end{captionabove}
\end{figure}

\subsection{Orientation}
\label{subsec:repOrientation}
Orientation is also a very interesting field for spatial representation and reasoning. Relative orientation of spatial entities is a ternary relationship depending on the referent, the relatum and the frame of reference (reference system) which can be specified either by a third object (origin) or by a given direction. The orientation is determined by the direction in which the referent is located in relation to the frame of reference.\par 

Three reference systems can be distinguished:
\begin{itemize}
	\item In the \textbf{Intrinsic Reference System }the orientation is  determined by an inherent feature of the relatum.
  \item The \textbf{Absolute Reference System} (also called extrinsic reference system) uses external parameters like global coordinate systems to give the relatum an orientation.  
  \item A third object (origin) is used in the \textbf{Relative Reference System} (also called deictic reference system) so that the orientation is defined by the "point of view" this third object has to the reference system.
\end{itemize}

Most spatial representations use points in a two-dimensional space as basic spatial entities since developing spatial orientations between extended objects is much more difficult. 

\subsection{Distance}
\label{subsec:repDistance}
The scalar entity distance is, together with topology and orientation, one of the most important aspects of space. In human communication qualitative descriptions of distance are used. There are absolute distance relations that specify distances between two objects like "close to", "far" or "very far". There also are relative distance relations that compare the distance between two objects with the distance to a third object like "closer than" or "further than". Qualitatively described absolute distances depend on the relative scale of space (for example objects on a desktop, in a home or in a city).\par

Combining distance relations not only depends on distance but also on orientation. If an object A is far from B and another object C is also far from B we cannot say anything certain about the distance between A and C. If the objects were on a line in the sequence of A B C, A would be very far from C, but if they were on a line in the order of A C B then A could be close to C. It is thus appropriate to use distance in combination with orientation which is called positional information.   

\subsection{Positional information}
In this thesis I'll use the term "positional information" for information structures that combine the orientation and distance representation. That means, that the position of an object is described by a combination of a qualitative orientation and a qualitative distance. Usually the number of possible relations (atomic relations) is the number of possible relations of the orientation representation multiplied with the number of relations of the distance representation. This may not be true for special cases, for example if relations are considered where objects are on the same location.\par
Hernández \cite{hernandez1994} on contrast uses the term "positional information" for combinations of orientation information with topological information.

\subsection{Orientated objects with extension}
All real world objects have an extension. If spatial representations would avail this fact improvements could be achieved for some applications. The extension is usually described by size and shape of the objects. The shape of an object could affect the frame of reference of an relation - in an intrinsic reference system it determines the orientation of this object. The size on the other hand has influence on how humans describe spatial relations. They prefer large, salient objects as reference objects: Humans rather say "The bicycle is in front of the cathedral" than "The cathedral is behind the bicycle".\par
Problems with extended objects arise if they overlap. Two orientations could then describe the same configuration. 
\cite{hernandez1994}

\subsection{Shape}

Shape is difficult to describe qualitatively. Using topology one can say whether an object has holes or whether it is in one piece or not. For finer grained distinctions shape primitives could be used. Other approaches characterize the boundary of an object using a sequence of different types of boundary segments or curvature extrema. Furthermore shape could be described with polygons that qualitatively define for each corner whether it is convex or concave, whether it is obtuse or right-angled or acute and with an qualitative description of the direction of the corner. Shape representations can make statements solely about the boundary or about the interior of an object \cite{Cohn2001}, \cite{Jungert}.

\subsection{Motion}
Motion is a spatio-temporal information that can be quite easily expressed qualitatively. An example for such an qualitative representation is given in \cite{brauer-qualitative}. In this approach, the motion of an object is observed from an deictic point of view with a fixed frame rate. The two components orientation and distance, that establish a vector, are used to describe the change of position of the object from one frame to the next. If the object is standing still the orientation and distance are 0. If two subsequent qualitative motion vectors are equal one can increment an index. A motion could be for example (distance,orientation):\\ $($close,forward$)^5$,$($close,left$)^9$,$($0,0$)^6$,$($far,forward$)^5$.\par
Another method of representation is one with an intrinsic frame of reference given by the orientation of the previous vector (the first vector is always "forward"). Now the frame rate is only used for standstills, in other cases a new qualitative value, the velocity, is used. A motion would now look like this:\\ $($short-dist,forward,slow$)$,$($medium-dist,left,slow$)$,$($0,0$)^6$,$($far,right,fast$)$.\par
A quite abstract representation is the propositional one. It identifies a set of movement shapes like "straight-line", "left-turn", "right-turn", "u-turn", "loop", etc. that can have relations between them that describe differences in magnitude, orientation and velocity.\cite{brauer-qualitative}


%
%
%
%
\section{Spatial Reasoning}
\label{sec:spatialReasoning}
To make use of their \sk, most applications have to use reasoning mechanisms to derive the knowledge needed. Respectively, in order for a representation to be useful, we most often have to consider not only its constituents and how they correspond to what is being represented, but also the operations on them. Only if simple search for matching relations is sufficient, one can disregard spatial reasoning. This is for example the case, if a robot perceives a configuration of objects and has to search for an object a user specifies by verbally describing an object configuration (like "go to the box that is right of the basket").\\
Other applications like navigation or computer vision that are presented in section \ref{sec:applications} make serious use of reasoning algorithms. Spatial reasoning is not possible without the presence of a spatial representation and the quality of the results the reasoning generates depends heavily on the underlying representation.\\
To deal with commonsense knowledge is important for any intelligent system and it has been early recognized as one of the central topics of \AI~ to represent and reason about commonsense knowledge. The question can be raised if the formal logic or the physical space is more fundamental for reasoning processes, \cite{freksa1992}. The aim of research on reasoning is to develop efficient algorithms, since the brute-force "generate-and-test" approach is always available but most often unusable due to it's complexity.\par
Reasoning can perform many tasks. Most importantly it can infer knowledge that is implicit in the knowledge base and make it thus explicit. That way the knowledge base is extended. Reasoning can also answer queries that are given with only partial knowledge and with a specific context. Various types of consistency can be maintained using reasoning techniques that simplify the use of the knowledge base. In general reasoning is used to acquire and process new knowledge.\par



\subsection{Algebraic structure}

In this section some basic reasoning techniques over spatial representations are presented. In order to do so, an assumption is made which actually holds for many calculi, especially for the calculi relevant for this thesis. The assumption is, that the relations between the objects that are being manipulated are ternary. Ternary relations are being made over three objects: the first argument is the origin, the second one is the relatum and the third one the referent. The relation describes the referent with respect to the frame of reference (or reference system) determined by the origin and the relatum.\par
In qualitative representations each representation has a finite number of atomic relations corresponding to the finite number of spatial configurations that are distinguished by the representation. The relation over the three objects corresponds to the spatial configuration of those objects. Special cases usually occur when two of the three the objects are on the same position. In order for the algebra to recognize those cases they are added to the set of atomic relations.\par
In general the three objects partition the space and the set of atomic relations is given by this partition. \\

\textbf{General relation} A (general) relation is any subset of the set of all atomic relations. The interpretation of such a relation $R$ is as follows: \\$\left(\forall X,Y,Z\right)\left(R\left(X,Y,Z\right) \Leftrightarrow \vee_{r\in R}r\left(X,Y,Z\right)\right)$.\\
The atomic relations are designed to be Jointly Exhaustive and Pairwise Disjoint (JEPD). That means, given any three objects $X,Y,Z$, there exists one and only one atomic relation $r$ such that $r\left(X,Y,Z\right)$.
\subsubsection{Unary operations}
The ternary relations have three arguments and thus there are 3! = 6 possible permutations for the arrangement of those arguments. Following \cite{zimmfre1996} the terminology shown in table \ref{tab:transformations} is used for those transformations.\par

\begin{table}
\caption{Transformations}
\label{tab:transformations}
\begin{tabular}{l|l|l}
term				&	symbol				& permutation \\\hline
identical		& \scshape{Id}	& $(\forall X,Y,Z)(R(X,Y,Z)\Rightarrow$ \scshape{Id} $(R(X,Y,Z)))$\\
inversion		& \scshape{Inv} & $(\forall X,Y,Z)(R(X,Y,Z)\Rightarrow$ \scshape{Inv} $(R(Y,X,Z)))$\\
short cut		& \scshape{Sc}  & $(\forall X,Y,Z)(R(X,Y,Z)\Rightarrow$ \scshape{Sc} $(R(X,Z,Y)))$\\
inverse short cut		& \scshape{Sci} & $(\forall X,Y,Z)(R(X,Y,Z)\Rightarrow$ \scshape{Sci} $(R(Z,X,Y)))$\\
homing			& \scshape{Hm}  & $(\forall X,Y,Z)(R(X,Y,Z)\Rightarrow$ \scshape{Hm} $(R(Y,Z,X)))$\\
inverse homing	&\scshape{Hmi}& $(\forall X,Y,Z)(R(X,Y,Z)\Rightarrow$ \scshape{Hmi} $(R(Z,Y,X)))$\\
\end{tabular}
\end{table}
\par


\subsubsection{Binary operations}
\textbf{Composition}

The composition $R_1 \otimes R_2$ of two relations $R_1$ and $R_2$ is the most specific relation $R_j$ such that:\\ $\left(\forall X,Y,Z,W\right)\left(R_1\left(X,Y,Z\right) \wedge R_2\left(X,Z,W\right) \Rightarrow R_j\left(X,Y,W\right)\right)$.\\ Given four objects $X,Y,Z,W$ and two atomic relations $r_1$ and $r_2$ the conjunction $r_1\left(X,Y,Z\right) \wedge r_2\left(X,Z,W\right)$ is trivially inconsistent if either of the following holds:\\
$\left(\forall X,Y,Z,W\right)\left(\left(r_1\left(X,Y,Z\right) \Rightarrow \left( Z = X \right)\right) \wedge
\left(r_2\left(X,Z,W\right) \Rightarrow \left( Z \neq X \right)\right)\right)$ \\
$\left(\forall X,Y,Z,W\right)\left(\left(r_1\left(X,Y,Z\right) \Rightarrow \left( Z \neq X \right)\right) \wedge
\left(r_2\left(X,Z,W\right) \Rightarrow \left( Z = X \right)\right)\right)$ \\
For atomic relations for which the above holds the result of the composition is empty: $r_1 \otimes r_2 = \emptyset$\par

In general a compositional inference is a deduction from two relational facts of the form $R_1(a,b)$ and $R_2(a,b)$ to a relational fact of the form $R_3(a,c)$, involving only a and c. The validity of compositional inferences does, in many cases, not depend on the specific elements involved but only on the logical properties on the relations. In such a case the composition of pairs of relations can be abstracted by table look up as and when required. Given the set of n JEPD atomic relations, one can store in a $n \times n$ composition table the relationships between $x$ and $z$ for a pair of relations $R_1(x,y)$ and $R_2(y,z)$. In general, each entry can be a disjunction of the base relations.\par
For the ternary relations those tables have to be made available for the following two cases:\\
$\left(\forall X,Y,Z,W\right)\left(\left(r_1\left(X,Y,Z\right) \Rightarrow \left( Z = X \right)\right) \wedge
\left(r_2\left(X,Z,W\right) \Rightarrow \left( Z = X \right)\right)\right)$ and\\
$\left(\forall X,Y,Z,W\right)\left(\left(r_1\left(X,Y,Z\right) \Rightarrow \left( Z \neq X \right)\right) \wedge
\left(r_2\left(X,Z,W\right) \Rightarrow \left( Z \neq X \right)\right)\right)$ \par
For a representation using a limited set of binary relations, the simplicity of the compositional inference makes it an attractive means of effective reasoning.\par

\textbf{Intersection} The intersection of two relations $R_1$ and $R_2$ is the relation $R$ which consists of the set-theoretic intersection of the atomic relation sets from $R_1$ and $R_2$ : $R = R_1 \cap R_2$.\par
\cite{moratz1999}

\subsection{Constraint Based Reasoning}
\label{sec:CSP}
The constraint satisfaction problem is identified as an abstract formulation of many difficult problems in \AI.
Given that qualitative representations can be expressed in form of relations, general constraint satisfaction techniques can be applied for various kinds of inference. Constraint satisfaction techniques play an important role in Computer Science as a whole, and in \AI~in particular. Many difficult problems involving search from areas such as (robot) navigation, temporal reasoning, graph algorithms and machine design and manufacturing can be considered to be special cases of the constraint satisfaction problem (CSP).\par

Constraint Satisfaction Problems (CSP) generally consist of:\\

\begin{itemize}
	\item a set of variables $X=\{X_1,...,X_n\}$, \\
	\item a discrete domain for each variable $\{D_1,...,D_n\}$ so that each variable $X_i$ has a finite set $D_i$ of 
possible values,\\
	\item a set $\{R_k\}$ of constraints, defined over some subset of the variable domains,
		$R_j \subseteq D_{i1} \times \cdots \times D_{ij}$, and showing the mutually compatible 
		values for a variable subset $\{X_{i1},...,X_{ij}\}$. Those constraints are restricting the values the variables can simultaneously take.
\end{itemize}

The problem is to find an assignment of values to variables such that all constraints are satisfied. Variants of the problem are to find all such assignments, the best one, if there exists any at all, etc. 

The constraints are restricted to be unary or binary because those are the operations we defined for the spatial representation. Furthermore it is assumed that all variable domains have the same cardinality.

The straightforward approach to find a satisfying assignment is a backtracking algorithm that corresponds to an uninformed systematic search. If, after having instantiated all variables relevant to a set of constraints, any of them is not valid, the algorithm backtracks to the most recently instantiated variable that still has untried values available. The run-time complexity of this algorithm is exponential, making it useless for realistic input sizes. This inefficiency arises because the same computations are repeated unnecessarily many times.\\
Possible solutions to this problem are the modification of the search space, the use of heuristics to guide the search or the use of the particular problem structure to orient the search. The latter one is especially interesting for spatial reasoning as we will see later.\par

\subsection{Consistency improvement}
The goal of modifying the search space is to avoid useless computation without missing any of the solutions of the original space. In other words, we are looking for a smaller but equivalent search space. This can be achieved either prior the search (by improving the consistency of the network through constraint propagation) or, in hybrid algorithms, during the search.\\

One way of reducing the number of repeated computations is constraint propagation. There those values are removed  from a domain that do not satisfy the corresponding unary predicates, as well as those values for which no matching value can be found in the adjacent domains such that the corresponding binary predicates are satisfied. The former process achieves node consistency, the latter arc consistency. Expressed differently one can say that this process of constraint relaxation is triggered by incompatible constraints. This concept of local consistency can be generalized to any number of variables. A set of variables is $k$-consistent if for each set of $k-1$ variables with satisfying values, it is possible to find a value for the $k$th variable such that all constraints among the $k$ variables are satisfied. A set of variables is strong $k$-consistent if it is $j$-consistent for all $j\leq k$. Of special interest is strong 3-consistency, which is equivalent to arc consistency plus path consistency. A network is path consistent if any value pair permitted in $R_{ij}$ is also allowed by any other path from $i$ to $j$. The process of achieving consistency in a network of constraints is called constraint propagation. Several authors give different definitions of constraint propagation, whose equivalence is not obvious.\par
One such definition is that constraint propagation is a way of deriving stronger (i.e., more restrictive) constraints by analyzing sets of variables and their related constraints. The value elimination consistency procedures mentioned above assume a very general extensional form of constraints as sets of satisfying value pairs. Whenever a value of the domain of a variable involved in more than one constraint is removed, a satisfying value pair might have to be removed from one or more of the other constraints, thus making them more restrictive.\par

Another view of constraint propagation is as the process of making implicit constraints explicit, where implicit constraints are those not recorded directly in compatible value pairs, but implied by them. This, however, can be seen as a side effect of the consistency procedures. The universal constraint, that allows any value pair, holds implicitly between two variables not explicitly linked together. If the domains of the two variables $X_i$ and $X_j$ are restricted to a few values, then the implicit constraint $R_{ij}$ can be made explicit as the set of combinations of the two domains. \par

Note that the global full constraint propagation is equivalent to finding the minimal graph of the CSP, where each value permitted by any explicit constraint belongs to at least one problem solution. Full constraint propagation is thus as hard as the CSP itself (in general NP-complete). The local constraint propagation techniques used to achieve node-, arc- and path-consistency have polynomial complexities and can be used as pre-processors that substantially reduce the need to backtrack during the search for a global solution.\par
Unfortunately, arc- and path-consistency do not eliminate in general the need to backtrack during the search because constraints are propagated only locally. 





\subsection{Spatial reasoning}

Spatial inferencing is concerned with the qualitative spatial analysis of the spatial configurations, often in two-dimensional Euclidean space. It uses the reasoning techniques presented in the previous sections. The composition of spatial relations is very under-determinate. The resulting relation sets tend to contain too many atomic relations. But often the structure of the spatial domain inherits certain constraints that allow further improvements for constraint based reasoning.\par
In section \ref{subsec:templogic} it is shown how Freksas approach uses constraints of the temporal domain to improve reasoning about time. An example for these constrains in the spatial domain is, that distinct solid objects never occupy the same point in space. Hernández \cite{hernandez1994}, giving another example, uses "abstract maps" that contain for each object in a scene a data structure with the same neighborhood structure as the domain required for the task at hand. A change in of the point of view can then be easily accomplished diagrammatically by rotating the labels of the orientation. Hernández also observed, that the composition of pairs of topological and orientation information yields more specific results.\par

In \cite{Freksa1991} Freksa shows, that spatial location representations of physical objects consist of connected parts (their "neighbors", see section \ref{sec:qsr}) and that movement in space is only possible between neighboring locations. Therefore a general representation should not explicitly check if these constraints hold; the constraints should rather be "build-in" in the representation and reasoning system. 


%
%
%
%
%
\section{Qualitative spatial calculi}
\label{sec:qualitative_spatial_calculi}


Qualitativeness can be best explained in contrast with it's counterpart Quantitativeness. First we gonna take a look at the American Heritage Dictionary entries for these two terms:
\begin{itemize}
\item quantity: 1. A specified or indefinite number or amount. An exact amount or number. 2. The measurable, countable, or comparable property or aspect of a thing.
\item quality: 1. The essential character of something; nature. 2. An inherent or distinguishing characteristic; property.
\end{itemize}

The most important words for quantity are number and measurable, because spatial quantities are measured which implies that a number is assigned to represent a magnitude. Usually the assignment can be made by a simple comparison. The magnitude of the quantity is compared to a standard quantity, the magnitude of which is arbitrarily chosen to have the measure 1.\par
The term quality is more difficult to explain. The qualitative representations can be characterized by establishing a correspondence between the abstract entities in the representation and the actual magnitudes. Quantitative knowledge is obtained whenever a standardized scale is used for anchoring the represented magnitudes. The use of a scale is also the context in which issues of granularity and resolution are meaningful, since a scale defines a smallest unit of possible distinction below which we are not able to say anything about a quantity.\\
\cite{brauer-qualitative}

Qualitative representation provides mechanisms for representing only those features that are unique or essential, whereas a quantitative representation allows to represent all those values that can be expressed with respect to a predefined unit. Although qualitative reasoning allows inferences to be made in absence of complete knowledge, it is not a probabilistic or fuzzy approach but it refuses to differentiate between certain quantities.\par

One primary goal of qualitative spatial representations is to provide a general vocabulary for performing efficient symbolic spatial inferences and analysis. Although the semantics of the vocabulary may not be unique, the discrete values for its definition over the continuous domains should allow computationally efficient spatial inference and less ambiguous spatial descriptions (i.e., with sufficient precision).\par

A mathematical definition for a qualitative abstraction can look like this:\\ 
Let $x$ be a quantitative variable, such that $x \in R$, and $R \subseteq \Re$. If the entire domain $R$ is partioned into a finite set of mutually disjoint subdomains $\left\{Q_1,Q_2,...,Q_m\right\}$, i.e., $\bigcup^{m}_{i=1}Q_i = R$, and, furthermore, all numerical values lying within $Q_i$ are treated as being equivalent and named symbolically by \textit{Label}$(Q_i)$, then the qualitative variable $[x]$ corresponding to $x$ is defined as follows:\par
$[x] \in X,\; X \subseteq \bigcup\limits^{m}_{i=1} \mbox{\textit{Label}}(Q_i)$\par
where \textit{Label}$(Q_i)$ is called a primitive qualitative value.\par
\cite{planningbuch}

\subsection{Properties of qualitative representations}
\label{sec:propertiesOfQSR}

This section is a citation from \cite{brauer-qualitative} who pointed out many of the useful properties of qualitative representations perfectly. 
\begin{itemize}
\item Qualitative representations make only as many distinctions as necessary to identify objects, events, situations, etc. in a given context (recognition task) as opposed to those needed to fully reconstruct a situation (reconstruction task).
\item All knowledge about the physical world in general, and space in particular, is based on comparisons between magnitudes. As representations that capture such comparisons, qualitative representations reflect the relative arrangement of magnitudes, but not absolute information about magnitudes. 
\item The search for distinctive features that characterizes the qualitative approach has an important side effect: It structures the domain according to the particular viewpoint used. Some of the qualitative distinctions being made are conceptually closer to each other than others. This structure is reflected in the set of relations used to represent the domain. 
\item Qualitative representations are "under-determined" in the sense that they might correspond to many "real" situations. The reason they still can be effectively used to solve spatial problems is that those problems are always embedded in a particular context. The context, which for simplicity can be taken to be a set of objects, should constrain the relative information enough to allow spatial reasoning, for example by making it possible to find a unique order along a descriptional dimension. In other words, a representation that can count on being used together with some particular context does not need to contain as much specific information itself. 
\item Qualitative representations handle vague knowledge by using coarse granularity levels, which avoid having to commit to specific values on a given dimension. With other words, the inherent "under-determination" of the representation absorbs the vagueness of our knowledge.
\item In qualitative representations of space, the structural similarity between the representing and the represented world prevents us from violating constraints corresponding to basic properties of the represented world, which in propositional systems would have to be restored through revision mechanisms at great cost. 
\item Unlike quantitative representations, which require a scale to be fixed before measurements can take place, qualitative representations are independent of fixed granularities. The qualitative distinctions made may correspond to finer or coarser differences in the represented world, depending on the granularity of the knowledge available and the actual context. 
\item The informative content of qualitative relations varies. Some describe what would correspond to a large range of quantitative values of the same quality, while others may single out a unique distinctive value.
\item While the discrimination power of single qualitative relations is kept intentionally low, the interaction of several relations can lead to arbitrarily fine distinctions. If each relation is considered to represent a set of possible values, the intersections of those sets correspond to elements that satisfy all constraints simultaneously. 
\end{itemize}
\cite{brauer-qualitative}


\subsection{Reasons for qualitative approaches}

Of course quantitative representations are very useful for many applications (CAD, 3D Graphics etc), but there are good reasons why quantitative approaches are to be preferred in many other systems:

\begin{itemize}
\item Advantages in Input and Output
\begin{itemize}
\item Partial and uncertain information: A value may not be exactly known (determined by the priori fixed scale) due to imprecise sensor data or vague human descriptions. It then either has to be ignored or assigned to a range of possible values in quantitative approaches. In the latter case there will be more computation needed. Qualitative approaches have no problems handling such information. 
\item Transformation: The transformation of a quantitative value to a qualitative one is done more easily and exactly than in the opposite direction. 
\item Qualitative input: The input for reasoning processes is often qualitative - it is often the result of a comparison rather than a quantitative description. It is thus better to use qualitative reasoning.
\item Real world input: Spatial reasoning (in the real world) is in most cases driven by qualitative abstractions rather than by complete a priori quantitative knowledge. 
\item Reasoning goal: The goal of reasoning is always qualitative. A decision is being processed not a quantitative value. 
\end{itemize}

\item Interaction with humans
\begin{itemize}
\item Missing adequacy: Humans are not very good at determining exact length, orientations ect. but they can easily perform context-dependent comparisons. In quantitative approaches humans would be forced to use quantities to express facts.  
\item Human reasoning: Humans also do qualitative reasoning more easily (and sometimes better).
\item Communication: Humans are used to communicate spatial facts qualitatively. This is thus their preferred way to interact with computers, too.
\item Human cognition: Qualitative representations are more transparent and intuitive to humans because it is believed that this is the method humans themselves use to reason about space. 
\end{itemize}

\item Computation issues
\begin{itemize}
\item Transformational impedance: As a consequence of the missing adequacy described above, spatial reasoning systems based on quantitative values might have to transform back and forth between their internal representation and the qualitative one which is used to communicated with humans. Information might get lost during these transformations.
\item Robustness: Qualitative approaches are more robust against errors than numerical methods. 
\item Falsifying effects: Quantitative models might falsify the representation by forcing discrete decisions. 
\item Unnecessary details: With qualitative representations unnecessary details can be omitted which yields to smaller and more transparent representations.
\item Properties: Quantitative approaches do not have the nice properties that their analytical counterparts have. 
\end{itemize}

\item Complexity
\begin{itemize}
\item Cheapness: Qualitative knowledge is cheaper than quantitative knowledge because it is less informative in a certain sense.
\item Complexity: The number of values that a descriptional value may take affects indirectly the complexity of the algorithms operating on them. Not only are more involved computations required, but the granularity of the representation, as determined by the fixed scale chosen, may make more distinctions than necessary for a given task. For example, the exact positions of all objects in the room are not necessary, if all we need to know is what objects are at a wall that is to be painted.
\end{itemize}
\end{itemize}

\cite{freksa1992} \cite{hernandez1994}




\subsection{Qualitative spatial representation}
\label{sec:qsr}

Quantitative representations usually store the \si~ in a common global or local coordinate system. Agents might have a local coordinate system derived from the inherent orientation of the agent or, in contrast, in the global coordinate system all entities and objects use the same common system which can, for example, be oriented on a building's floor plan or the cardinal directions. Agents can communicate their \sk~ by mapping their local coordinate systems to a global one. That can only be done if all agents have access to the global coordinate system. The relevant global coordinates can then be exchanged.\par

Using quantitative representations, it is difficult to handle indeterminate or inexact knowledge. For a quantitative representation the precise position and the size of all objects must be known. Furthermore, detecting certain spatial configurations is quite difficult but it is often needed to trigger certain actions or behaviors. In the numerical approach reasoning is done with numerical or geometrical methods like computing a tangent to an object or closest points of two objects to each other.\cite{qsrrenz}\par

The qualitative approach is representing \sk~without exact numerical values. It uses a finite vocabulary that describes a finite number of possible relationships. This is, compared to the quantitative approach, closer to how humans represent \sk. In natural language the human says something like "A is above B" or "A is next to B" to express spatial information. Those information are usually sufficient to identify an object or follow a route. It is easy to represent indefinite or uncertain knowledge with qualitative methods. For example, one could say "A is left or behind B". Furthermore are rules easy to define "Do something if A is in front of B".\par

Qualitative spatial knowledge is not inexact, even though no numerical values are used. This is because distinctions are only made if necessary and they depend on the level of granularity that was chosen. Both, the quantitative approach and the qualitative approach, have their own right because both have applications where they are best suited. Robots that interact with humans usually take advantage of using a qualitative representation. If available, exact coordinate based knowledge should be used. The parallel use and transformation of both representation approaches is often useful, for example for a car navigation system by giving directions that guide the way in an unknown city using its street map and data from the Global Positioning System. \par

\subsubsection{Points versus areas}  
Qualitative spatial knowledge can be represented using spatially extended objects or abstract points. For the one dimensional time Allen \cite{Allen} pointed out good reasons why to use intervals of time instead of points. He showed that every event in time (for example "finding a letter") can be decomposited into a time interval ("looking at a letter" -> "realizing that it is the one that is searched for"). This also holds true for space since every real life point can be magnified to an area (at least as one stays above nano-sized structures). Another reason why Allen preferred intervals of time is the problem of open or closed intervals that occurs if one models intervals of time with points of time. Considering a situation where a light is switched off there is a point in time where the light is neither on or off assuming open intervals. With closed intervals there would be a point in time where the light is both - on and off. Allan solved there problems very elegantly with his Temporal logic \ref{subsec:templogic}.\par

In two dimensional space Allens intervals corresponds to the aspects topology and shape. It is clear that it is needed to consider areas and not points if one wants to reason about topology or shape. But problems arise in two dimensional space because there are lots of possible classes of shapes which cannot be handled equally well. Freksa \cite{freksa1992} showed that it is more convenient to use points, especially for other aspects of space like orientation and distance. One reason is, that the properties of points and their spatial relations hold for the entire spatial domain. Another, that shapes can be represented using points at different levels of abstraction. Lastly Freksa explained, that it is desirable to be flexible with respect to the spatial entities and their resolution. That means, that in one context one might be purely interested in 0-dimensional points such as points on a map. Other applications might be interested in 1-dimensional information like the width of a river or the length of a road. 2-dimensional projections (e.g. area of a lake) or 3-dimensional shapes of objects might be of interest in other contexts.

\subsubsection{Orientation}
\label{subsubsection:orientation}
\begin{figure}[ht]
  \begin{captionabove}{\textbf{a)} location of \emph{c} wrt. the location of \emph{b} and the orientation \emph{ab}; \textbf{b)} Orientation relations wrt. to the location \emph{b} and the orientation \emph{ab}; \textbf{c)} left/right and front/back dichotomies in an orientation system}
	  \label{fig:orientation}
		\begin{center} 	\includegraphics[height=5.5cm]{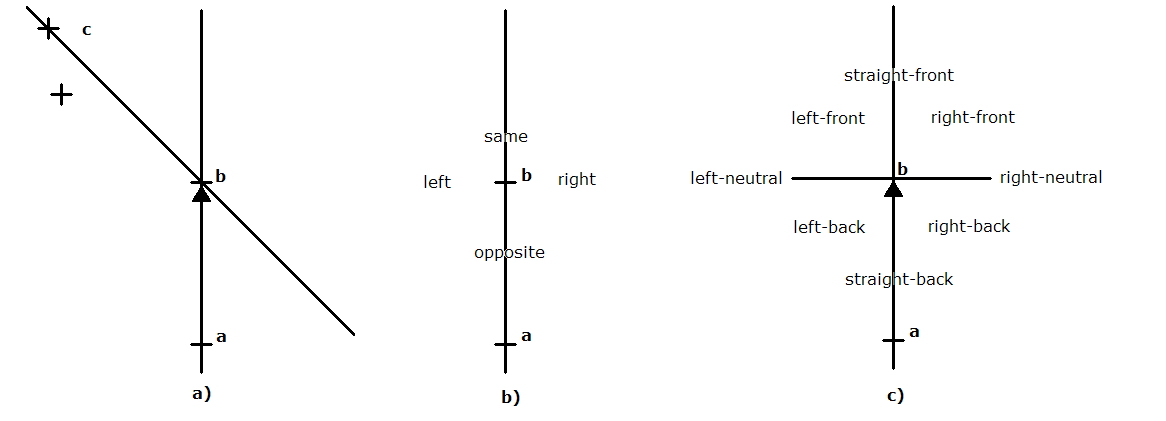} \end{center}
	\end{captionabove}
\end{figure}

Orientation is a ternary relationship depending on the referent, the relatum and the frame of reference which can be specified either by a third object (origin) or by a given direction (see section \ref{subsec:repOrientation}). If the frame of reference is given the orientation can be expressed using binary relationships. For 2-dimensional space Freksa \cite{freksa1992} defined orientation as a 1-dimensional feature which is determined by an oriented line. The oriented line is defined by an ordered set of two points.  
An orientation is then denoted by an oriented line \emph{ab} through the two points \emph{a} and \emph{b} (see figure \ref{fig:orientation}). \emph{ba} denotes the opposite orientation. The relative orientation in 2-dimensional space is then given by two oriented lines (which are represented by two ordered sets of points). The two ordered sets of points can share one point without loss of generality because the feature orientation is independent of location and vice versa. One can thus describe the orientation of a line \emph{bc} relative to the orientation line \emph{ab}. This way the ternary relationship is again achieved. Three special cases arise if the locations of (1) \emph{a} or (2) \emph{c} or (3) both are identical with the location of \emph{b}. In the first special case (\emph{a}=\emph{b}) no orientation information can be represented. In the second (\emph{c}=\emph{b}) and the third case (\emph{a}=\emph{b}=\emph{c}) orientation information is unavailable, too, but location information of \emph{c} is still available. The point \emph{c} is called the referent (also primary object or located object), \emph{b} the relatum (also reference object) and \emph{a} is called origin (also parent object).\par

From this point different approaches use diverse methods to determine the qualitative orientation values. A simple distinction can be made by defining four qualitatively different orientation values labeled \emph{same}, \emph{opposite}, \emph{left} and \emph{right}. If the point \emph{c} is on the line \emph{ab} on the other side of \emph{b} than \emph{a} the orientation is \emph{same} and if it is on the line \emph{ab} on the other side of \emph{a} than \emph{b} the orientation is \emph{opposite}. The orientation is \emph{left} if the point \emph{c} is located on the left semi-plane of the oriented line \emph{ab} and it is \emph{right} if \emph{c} is located on the right semi-plane of the oriented line \emph{ab}. Freksa observed that unlike in the case of linear dimensions, incrementing quantitative orientation leads back to previous orientations. In this sense, orientation is a circular dimension.\par
\begin{figure}[ht]
  \begin{captionabove}{cone-based representation}
	  \label{fig:cone}
		\begin{center} 	\includegraphics[height=7cm]{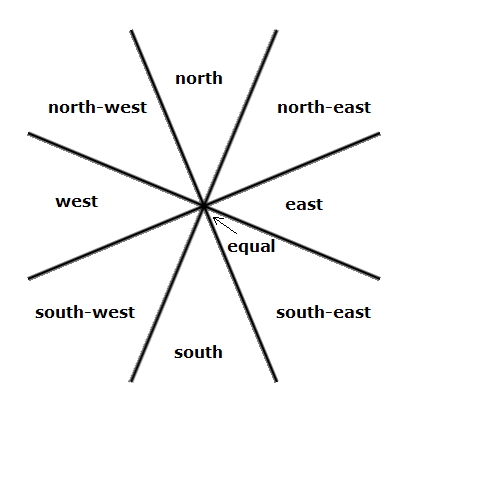} \end{center}
	\end{captionabove}
\end{figure}

A substantial gain of information can be achieved by introducing a front/ back dichotomy. In this so called \emph{cardinal direction representation} the eight different orientation labels \emph{straight-front}, \emph{right-front}, \emph{right-neutral}, \emph{right-back}, \emph{straight-back}, \emph{left-back}, \emph{left-neutral} and \emph{left-front} are distinguished (see figure \ref{fig:orientation}). Frank \cite{frank} called this method "projection-based" and Ligozat \cite{ligozat1998} "cardinal algebra". Ligozat also found that reasoning with this cardinal algebra is NP-complete. Another approach to augment the number of orientation values is called "cone-based" by Frank. Here nine different relations are distinguished: \emph{north}, \emph{north-east}, \emph{east}, \emph{south-east}, \emph{south}, \emph{south-west}, \emph{west}, \emph{north-west} and \emph{equal} (see figure \ref{fig:cone}). In this method the plane is split into eight slices of 45° by three lines. All eight segments have the same scope unlike in the cardinal algebra, where the \emph{-front} and \emph{-back} relations correspond to an infinite number of angles (as in the cone-based approach) while the \emph{straight-} and \emph{neutral-} relations correspond to a single angle.\par
Other approaches are presented more detailed in the chapter \ref{sec:StateOfTheArt} State of the Art.

\subsubsection{Distance}
\label{sec:qsrDistance}

Distance is, unlike topology and orientation, a scalar entity. As mentioned in section \ref{subsec:repDistance} one has to distinguish between absolute distance relations and relative distance relations. Absolute distance relations indicate the distance between two points and divide the real line into a different number of sectors depending on the chosen level of granularity. They can be represented quantitatively or qualitatively and depend on an uniform global scale. Absolute distance relations \textbf{name} distances whereas relative distance relations \textbf{compare} the distance of two points with the distance to a third point. Next to the obvious predicates <, =, > more relations for relative distances can be defined if needed (e.g. "much shorter", "a little bit less", "much longer").\par
In \cite{hernandez1997} a general framework for representing qualitative distances at different levels of granularity has been developed. The space around the relatum $RO$ is partitioned according to a number of totally ordered distance distinctions $Q = \{q_0,q_1,q_2,...,q_n\}$, where $q_0$ is the distance closest to the relatum and $q_n$ is the one farthest away (which can go to infinity). Distance relations are organized in \emph{distance systems} $D$ defined as:\par
$D=(Q,A,\Im)$\par
where:
\begin{itemize}
	\item $Q$ is the totally ordered set of distance relations;
	\item $A$ is an \emph{acceptance function} defined as $A : Q \times O \rightarrow I$, such that, given a reference object (relatum) $RO$ and a set of objects $O$, $A(q_i,RO)$ returns the geometric interval $\delta_i \in I$ corresponding to the distance relation $q_i$;
	\item $\Im$ is an algebraic structure with operations and order relations defined over a set of intervals $I$. $\Im$ defines the \emph{structure relations} between intervals.
\end{itemize}
Each distance relation can be associated to an \emph{acceptance area} surrounding a reference object (relatum) which will be circular in isotropic space (that is space which has the same cost of moving in all directions).\par

\emph{Homogeneous} distance systems are those in which all distance relations have the same structure relations. That means, the size of the geometric intervals $\delta_i$ follows a recurrent pattern. The general type of distance systems where this is not the case is called accordingly \emph{heterogeneous}. More restrictive properties of the structure of these intervals include \emph{monotonicity} (each interval is bigger or equal than its previous) and \emph{range restriction} (any given interval is bigger than the entire range from the origin to the previous interval).\par
The frame of reference (see section \ref{subsec:repOrientation}) is important for distance systems, too. In the \emph{intrinsic} reference system the distance is determined by some inherent characteristics of the relatum, like its topology, size or shape (e.g. 50 meters can be considered far away from a small house but they seem to be close if standing next to a skyscraper). The distance is determined by some external factor in the \emph{extrinsic} reference system, like the arrangement of objects, the traveling time or the costs involved.  The \emph{deictic} reference system uses an external point of view to determine the distance, like if the objects are visually perceived by an observer.

\subsubsection{Conceptual Neighbors}
\label{subsection:neighbors}


For higher-level reasoning and knowledge abstraction the conceptual neighborhood relation provides several advantages. It was originally developed by Freksa for temporal knowledge \cite{FreksaTemporal}, generalizing Allens temporal logic (see section \ref{subsec:templogic}). Later Freksa successfully applied the conceptual neighborhood principle to spatial knowledge \cite{freksa1992}.\par
In a representation two relations are conceptual neighbors, if there exists an operation in the represented domain that causes a direct transition from one relation to the other. Those operations can be either spatial movement or deformations for the physical space. In the cone-based approach described above each relation except \emph{equal} has three conceptual neighbors. The conceptual neighbors of \emph{north}, for example, are \emph{north-east}, \emph{north-west} and \emph{equal} because it is possible to make a direct move there without the need to traverse any other relation. \emph{East} on the other hand is not a conceptual neighbor of \emph{north} because there is no direct transition from \emph{north} to \emph{east} - possible ways either have to cross \emph{north-east}, \emph{equal} or even detour by traversing through \emph{north-west}. But this only holds for the cone-based representation presented here - other representations have other conceptual neighbors and for more coarse ones there might be a direct way from north to east. The conceptual neighbors for a distance relation $q_x$ in the distance representation above are obviously those two distance relations in the totally ordered set $Q$ that are next to $q_x$, viz $q_{x-1}$ and $q_{x+1}$ (there is only one neighbor for the first ($q_0$) and last).\par
A great benefit of conceptual neighborhood structures is, that they intrinsically reflect the structure of the represented real world with their operations. This makes it possible to implement reasoning strategies which are strongly biased toward the operations in the represented domain. Conceptual neighborhoods allow to only consider operations which are feasible in the specific domain which can restrict the problem space and thus achieve nice computational advantages.    


\subsection{Qualitative spatial reasoning}
\label{sec:qualitativespatialreasoning}

\begin{figure}[ht]
  \begin{captionabove}{The possible locations of \emph{c} (gray) before and after the unary \textsc{Inv} operation }
	  \label{fig:exampleunary}
\begin{center}
		\includegraphics[height=5.5cm]{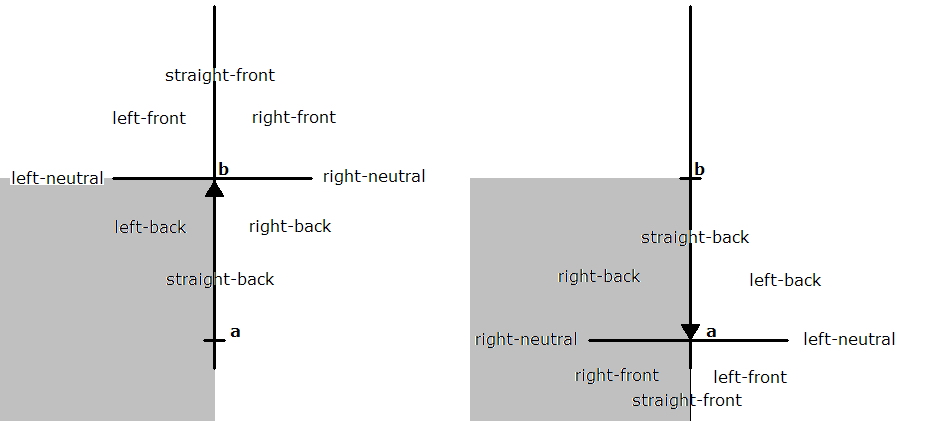}
\end{center}
	\end{captionabove}
\end{figure}

In order to use the reasoning techniques introduced in section \ref{sec:spatialReasoning} the unary and binary operations have to be defined for the spatial representation. In this section these operations are illustrated by using the cardinal direction representation presented in the orientation section above (see figure \ref{fig:orientation} \textbf{c)} ).\par
The first example will be the unary operation of a spatial arrangement where \emph{c}, the referent, is at the \emph{left-back} of \emph{ab}, where \emph{a} is the origin and \emph{b} the relatum. Now we might need to exchange the origin and the relatum. That is done by using the inverse (\textsc{Inv}) operation (see table \ref{tab:transformations}). \emph{B} is now the origin and \emph{a} the relatum. The result of that operation is ambiguous, as can be seen in figure \ref{fig:exampleunary}. The space is now partitioned differently because the front/ back dichotomy now divides the space through \emph{a}. Of course the labels changed, too. For example, what was left is now right and vice versa because the direction has been changed by 180°. Because of the qualitative nature of the representation we cannot say whether the point \emph{c} is \emph{right-back}, \emph{right-neutral} or \emph{right-front} of \emph{ba}. The result of this operation is thus a disjunction of all three possible atomic relations.\par

\begin{figure}[ht]
  \begin{captionabove}{Composition - the possible locations of \emph{d} are shown in gray}
	  \label{fig:examplecomp}
\begin{center}
		\includegraphics[height=7.66cm]{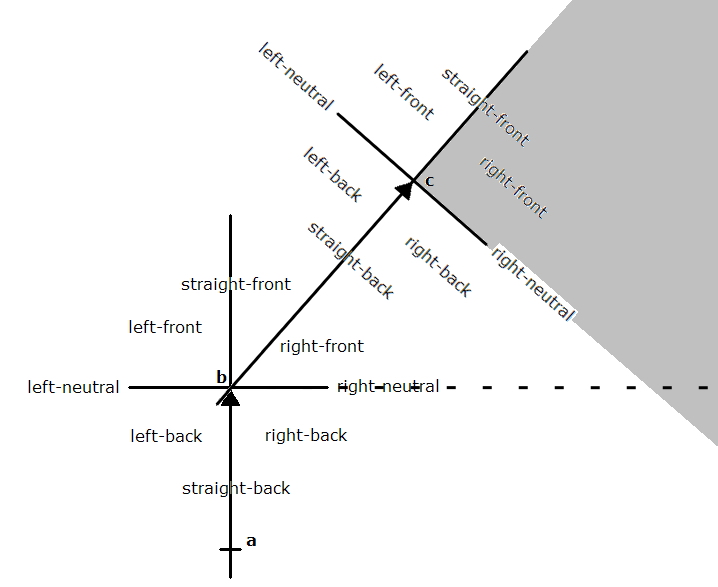}
\end{center}
	\end{captionabove}
\end{figure}

The binary composition operation can have ambiguous results as well. We now have four points \emph{a, b, c} and \emph{d} and two spatial relations - one between \emph{a, b,} and \emph{c} and the other between \emph{b, c} and \emph{d} (always origin, relatum and referent in this order). The composition of these two relations answers the question where the point \emph{d} is in relation to \emph{a} and \emph{b} (see figure \ref{fig:examplecomp} - the possible locations of \emph{d} (for an assumed position of \emph{c}) are indicated gray). In both input-relations the referent (\emph{c} respectively \emph{d}) is located in the \emph{right-front}. It is clearly visible that the composition result, which describes the location of \emph{d} (gray) with \emph{a} as origin and \emph{b} as relatum, consist of the atomic relations \emph{right-front}, \emph{right-neutral} and \emph{right-back}. The result of this composition is thus a disjunction of all three possible atomic relations.\par






%
%
%
%
%
\section{Applications of qualitative spatial reasoning}
\label{sec:applications}


Qualitative representations of space can be used in various application areas in which spatial knowledge plays a role. It is in particular used in those systems which are characterized by uncertainty and incompleteness.\\
This coarse knowledge often occurs in early stages of design projects, in which verbally expressed spatial requirements and raw sketches are very common. Examples for such application areas include spatial and geographical information systems, computer aided systems for architectural design and urban planning, document analysis, computer vision, natural language processing (input and output), visual (programming) languages, qualitative simulations of physical processes and of course navigation, robot-navigation and robot-control.\\
Although so many areas exists, there are quite few existing applications present. One reason for that is, that most applications require more than just one aspect of space. But since the different aspects of space are not independent of each other it is not possible to simply add different approaches covering single aspects of space to achieve the desired functionality. Another reason for the lack of existing applications is, that many approaches have no or not complete reasoning mechanisms without which the spatial representations are less useful \cite{qsrrenz}.\par

\subsubsection{Geographical Information Systems (GIS)}
GIS are these days commonplace, but they have major problems with interaction with the user. GISs have access to a vast size of vectorized information, without the ability to sufficiently support intuitive interaction with humans. Users may wish to perform queries that are essentially, or at least largely, qualitative. Egenhofers concept of \emph{Naive Geography} \cite{Egenhofer:95} employs qualitative reasoning techniques to enhance the human - GIS interface.
 
\subsubsection{Robotics}

Robotic navigation can make use of qualitative approaches in high level planning. Besides that, robots may navigate using qualitative spatial representations if the robot's model of its environment is imperfect, which would lead to an inability to use standard robot navigation techniques if numerical representations had been chosen. Research in qualitative methods for robotics concentrates in robust qualitative representations and inference machines for exploration, mapping and navigation.\par

In \cite{FreMo2000} an approach to high level interaction with autonomous robots by means of \textbf{schematic maps} was outlined. Schematic Maps are knowledge representation structures that encode qualitative spatial information about a physical environment. In the presented scenario an autonomous mobile robot with rudimentary sensory abilities to recognize the presence as well as certain distinguishing features of obstacles had the task to move to a given location within a structured dynamic spatial environment. In order to achieve that task the robot has to implement certain abilities: It has to determine where to go to reach the target location which needs spatial knowledge. It further has to compute what actions are to be made in order to move there. This needs knowledge about the relation between motor actions and movements and it also needs knowledge of the relation between movement and the robots spatial location.\\
In robotics there are no detailed, complete information about the spatial structure of the environment available for several reasons. First it is hard to provide detailed spatial knowledge which agrees with the physical space. Secondly there is no persistence - that means that the spatial configurations might change for unpredictable reasons. Lastly the actions of the robots in physical space are typically not fully predictable due to slipping wheels and imprecise sensors and motors.\\
One main reason why autonomous robotics is so difficult is, that the robot lives in two worlds simultaneously. It exists in the physical world of objects and space as well as in the abstract world of representation and computation. Those worlds are incommensurable: There is no theory that can treat both worlds in the same way.
Classical \AI~approaches try to develop formal theories about space that are sufficiently precise to describe all that is needed to perform the actions on the level of the representation (successful examples: board games). Qualitative theories on the other hand only deal with some aspects of the physical world and leave other aspects to be dealt with separately. \\
In robotics maps are used for communication between the robot and humans as well as interaction interfaces between the robot and its environment. The power of maps as representation media for spatial information stems from the strong correspondence between spatial relations in the map and spatial relations in the real world. Thus spatial relations may be directly read from the map, even if those have not been entered into the representation explicitly. Maps distort spatial relations to some extent, most obviously they transform the scale.\par
Schematic Maps distort beyond the distortions required for representational reasons to omit unnecessary details, to simplify shapes and structures, or to make the maps more readable. Typical examples for schematic maps are public transportation maps or tourist city maps which may severely distort distances and orientations between objects. 
Schematic maps provide suitable means for communicating navigation instructions to robots: spatial relationships like neighborhood, connectedness of places, location of obstacles, etc. can be represented. Those maps can be encoded in terms of qualitative spatial relations and qualitative spatial reasoning can be used to infer relationships needed for solving the navigation task. For the correspondence between the map and the spatial environment coarser low-resolution information is more suitable. A large number of rare or unique configurations can be found this way. If relations from the map and the environment do not match perfectly, the conceptual neighborhood knowledge (see section \ref{subsection:neighbors}) can be used to determine appropriate matches.\\
Schematic maps may be created in at least three different ways: A human may acquire knowledge about the spatial layout and build the schematic map by entering the relevant relationships, perhaps with the help of a computerized design tool. The robot itself may explore its environment in its idle time and create a schematic map that reflects landmarks and spatial relationships between notable entities discovered from the robots perspective. Furthermore a spatial data base could be used to create schematic maps.\par
The navigation planning and execution starts with the initial schematic map that provides the robot with survey knowledge about its environment. Important features are extracted from the map for identification in the environment. The robot can also enter discoveries into the map that it made during its own perceptual explorations. A coarse plan for the route is being produced using global knowledge from the map and local knowledge from its own perception. The resulting plan is a qualitative one. During its execution, the robot will change its local environment through locomotion. This enables it to instantiate the coarse plan by taking into account temporary or unpredicted obstacles. Also, the local exploration may unveil serious discrepancies between the map and the environment that prevent the instantiation of the plan. In this case, the map can be updated by the newly accumulated knowledge and a revised plan can be generated.

\subsubsection{Linguistics}
Humans naturally communicate using qualitative expressions. Thus humans can construct schematic maps rather easily. Therefore schematic maps can be used for two-way communication between humans and robots. The instructor can give a verbal description of the goal. If the robot fails to generate a working plan for some reason it can communicate with the instructor and possibly give an indication about it's (spatial) problem. However, in natural language, the use and interpretation of spatial propositions tend to be ambiguous. There are multiple ways in which natural language spatial relationships can be used.\\
\cite{FreMo2000}





\nocite{bkb2000}

\subsubsection{Voronoi graphs}

\begin{figure}[ht]
	\begin{captionabove}{The fine lines are the generalized Voronoi diagram of a 2D environment. On the right side is the corresponding GVG.}
		\label{fig:voronoi1}
\begin{center}
  	\includegraphics[width=9cm]{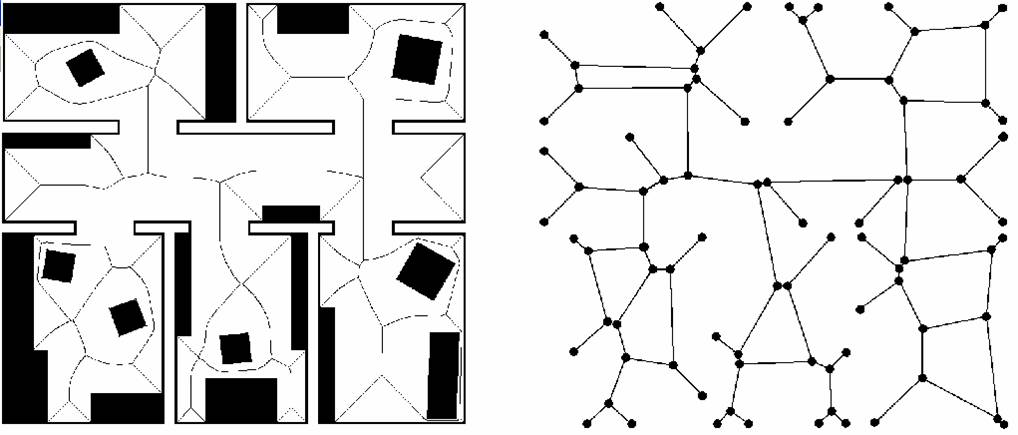}
\end{center}
	\end{captionabove}
\end{figure}

In \cite{cosy_Moratz_03_Spatial} Moratz and Wallgrün used the \doi~ presented in section \ref{sec:doi} to build an environmental map for mobile robot navigation. The foundation of this map is a generalized Voronoi graph (GVG), which is the graph corresponding to the generalized Voronoi diagram of the robots free space that is annotated with additional information (see figure \ref{fig:voronoi1}). The nodes of the GVG correspond to the meet points and the edges correspond to the Voronoi curves. This graph only represents the topology of the GVD and it is annotated with additional information, including DOI information between the neighboring vertices.\par
This representation brings together advantages stemming from the use of topological maps and from the use of Voronoi diagrams in mobile robot applications. Path planning can be done directly by using graph search to search through the GVG. Since only qualitatively different paths are represented path planning becomes very efficient. \cite{diplomWallgruenVoronoi} The DOI information stored in the graph can be used for cycle detection. The DOIs are propagated along the path through the GVG that connects the vertices that are checked to be identical.

%
%
%
%
%


%
%
%
%
%

%% file: chapters/stateOfArt.tex
%
%
%
%

\chapter{State of the Art}
\label{sec:StateOfTheArt}

%
%
%
%
%
\section{RCC-8}
\label{sec:rcc}

In 1992 Randell, Cui and Cohn developed the Region Connection Calculus \cite{Randell}. In this theory topological relations are used as a basis for qualitative spatial representation and reasoning. 

RCC-8 is a subset of eight relations from RCC. Those eight base relations are: "A is disconnected from B", "A is externally connected to B", "A partially overlaps B", "A is equal to B", "A is a tangential proper part of B", "A is a non-tangential proper part of B" and the converse of the latter two relations (A and B are spatial regions). RCC-8 also contains all possible unions of these base relations. RCC-8 is the spatial counterpart of Allen's temporal interval algebra - the eight base relations correspondent to Allen's base relations. Most other relations in RCC are refinements of the RCC-8 base relations. \par

RCC-8 semantics of base relations can be described using propositional logics rather than first-order logic needed in RCC \cite{Bennett}. Thus reasoning about TCC-8 relations is decidable. 

RCC is a fully axiomatized first-order theory for representing topological relations. All spatial entities are regarded as spatial regions. See appendix \ref{app:topology} for detailed definitions in the RCC calculus \cite{qsrrenz}.

%
%
%
%
%
\section{The Dipole Calculus}
\label{sec:dipole}
\begin{figure}[ht]
	\begin{captionabove}{Dipole}
  	\label{fig:dipol1}
\begin{center}
		\includegraphics[width=2.5cm]{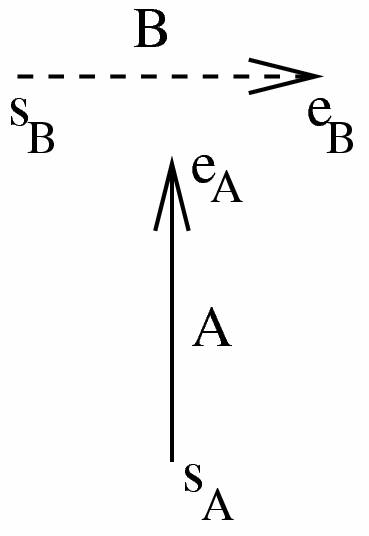}
\end{center}
	\end{captionabove}
\end{figure}

An approach for dealing with intrinsic orientation information is presented in \\\cite{dipol}. 
It uses orientated line segments called dipole that are formed by a pair of points - a start point ($s_A$ for a dipole A) and an end point ($e_A$ for a dipole A). These dipoles are used to represent two-dimensional extended spatial objects with an intrinsic orientation (see figure \ref{fig:dipol1}). The local orientation of the dipoles is even simpler as those relation presented in section \ref{subsubsection:orientation}. In the dipole representation a point can only be \emph{left} (l), \emph{right} (r) and \emph{on} the straight line (o) of the referring dipole, merging the atomic relations \emph{same} and \emph{opposite} of figure \ref{fig:orientation} b) into one. With the \emph{left} and \emph{right} relations between a dipole and the start and end points of another dipole, 24 JEPD atomic relations can be distinguished, given that no more than two points are allowed to be on a line (this is called general position) with the exception that two dipoles may share one point.\par

Given two dipoles A and B with their start points $s_A$, $s_B$ and their end points $e_A$, $e_B$, the atomic relations between them can be described as follows:\par
$A R_1 s_B \wedge A R_2 e_B \wedge B R_3 s_A \wedge B R_4 e_A$ with $R_1, R_2, R_3, R_4 \in \{r,l,s,e\}$\par

\begin{figure}[ht]
	\begin{captionabove}{The 24 atomic relations of the dipole calculus}
  	\label{fig:24dipoles}
\begin{center}
		\includegraphics[width=12cm]{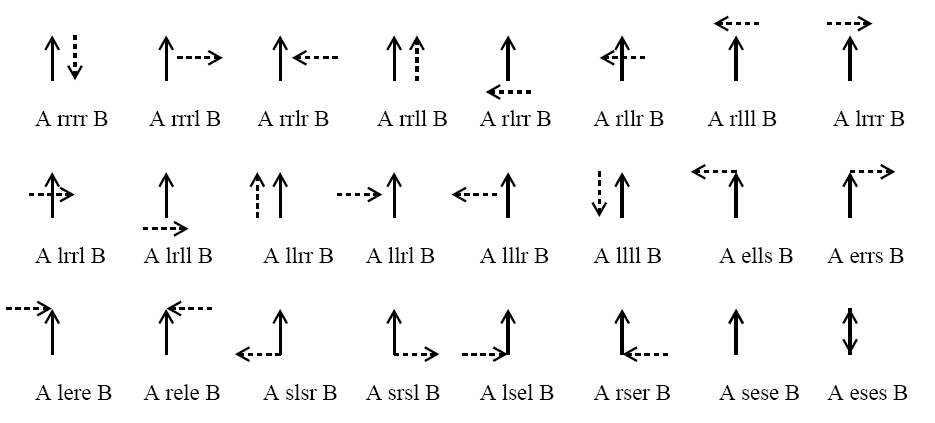}
\end{center}
	\end{captionabove}
\end{figure}

If $R_1$ is $s$ or $e$ (meaning that $s_B$ is on the same point as $s_A$($s$) or $e_A$($e$)), $R_2$ can only be $r$ or $l$ and vice versa ($R_1$ and $R_2$ exchanged), since dipoles share maximal one point. This also holds for $R_3$ and $R_4$.\\
A short form of the term above can be written as: $ A R_1R_2R_3R_4 B$ (see figure \ref{fig:24dipoles})

\subsection{Constraint based reasoning with the Dipole Calculus}
The Dipole Calculus forms a relation algebra - it is closed under the binary composition operation and under the unary intersection, complement and converse operation. It also supports an empty relation, an universal relation and an identity relation. Compound relations are sets of atomic relations. The composition results of atomic relations are obtained in an composition table. The composition of compound relations can be obtained as the union of the compositions of the corresponding atomic relations. Dipoles Constraint Satisfaction Problems can be solved using the methods presented in section \ref{sec:CSP}. It has been shown that the Dipole Calculus is NP-hard and in PSPACE.\par
Freksas double-cross calculus 
describes relations between triples of points, which can be regarded as relationships between a dipole and an isolated point. In contrast to Freksas ternary relations, the dipole relations are binary relations which makes reasoning much easier. Also, Freksa distinguishes more possible relations between a dipole and a point than the dipole.\\
A difficulty of the the Dipole Calculus is, that it presumes intrinsic objects although the intrinsic character of objects may not be existent or invisible for an application's sensor.

%
%
%
%
%



%
%
%
%
%
\section{\tpcc}
\label{sec:tpcc}
The \tpcc~ (TPCC) is a \qsr~ calculus that uses ternary relations of points developed by Reinhard Moratz\\  \cite{moratzTPCC}. The distinctions in TPCC are less coarse then in the calculi described above and thus TPCC permits more useful differentiations for realistic application scenarios.

\subsection{Flip-Flop}

In TPCC we again have point-like objects on a 2D-plane. A relative reference system is given by an origin and a relatum. The origin and the relatum define the reference axis. The spatial relation between the reference system and the referent is then described qualitatively by naming the part of the partition in which the referent lies. 
\begin{figure}[ht]
	\begin{captionabove}{The flip-flop partition}
		\label{fig:flip-flop}
\begin{center}
		\includegraphics[width=7cm]{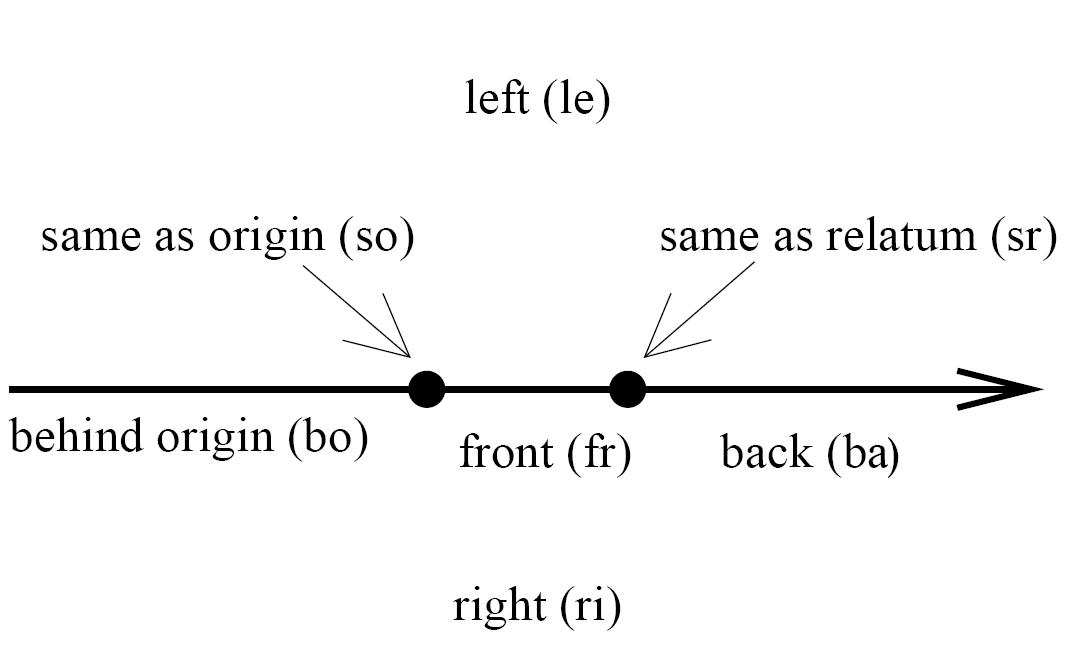}
\end{center}
	\end{captionabove}
\end{figure}

Ligozat \cite{Ligozat1993} suggested a system which he called flip-flop calculus. In this calculus the reference axis partitions the 2D-plane into two parts - left and right. The spatial relation between the reference system and the referent is described qualitatively by naming the part of the partition in which the referent lies. The referent can not only be on the left or the right - it can also be on the reference axis itself. In this case Ligozat distinguishes five configurations. The referent can either be behind the relatum (back - ba), at the same position as the relatum (same as relatum - sr), in front of the relatum (front - fr), at the same point at the origin (same as origin - so) or behind the origin (behind origin - bo). shows this partition. The partition is shown in figure \ref{fig:flip-flop}.

Points A, B and C can be examples for origin, relatum and referent. Isli and Moratz \cite{IsliMoratz1999} observed two additional configurations in which origin and relatum have exactly the same location. In the first configuration the origin and relatum are at the same point and the referent at some other (double point - dou). Secondly all three points can be at the same location (triple point - tri). 

Infix notation is used to describe configurations. The reference system consisting of origin and relatum are written in front of the relation symbol - the referent is written behind it.

Vorweg et al. \cite{vorweg} showed empirically that the acceptance regions for front and back need similar extensions like left and right. In the flip-flop calculus front and back only have linear acceptance regions. 

Freksa \cite{freksa1992} extended the flip-flop calculus by partitioning the 2D-plane with a cross. Therefore the left and right side is respectively divided into a front and back

\subsection{Representation in TPCC}

\begin{figure}[ht]
	\begin{captionabove}{Names of the TPCC areas}
  	\label{fig:TPCCnames}
\begin{center}
		\includegraphics[width=5cm]{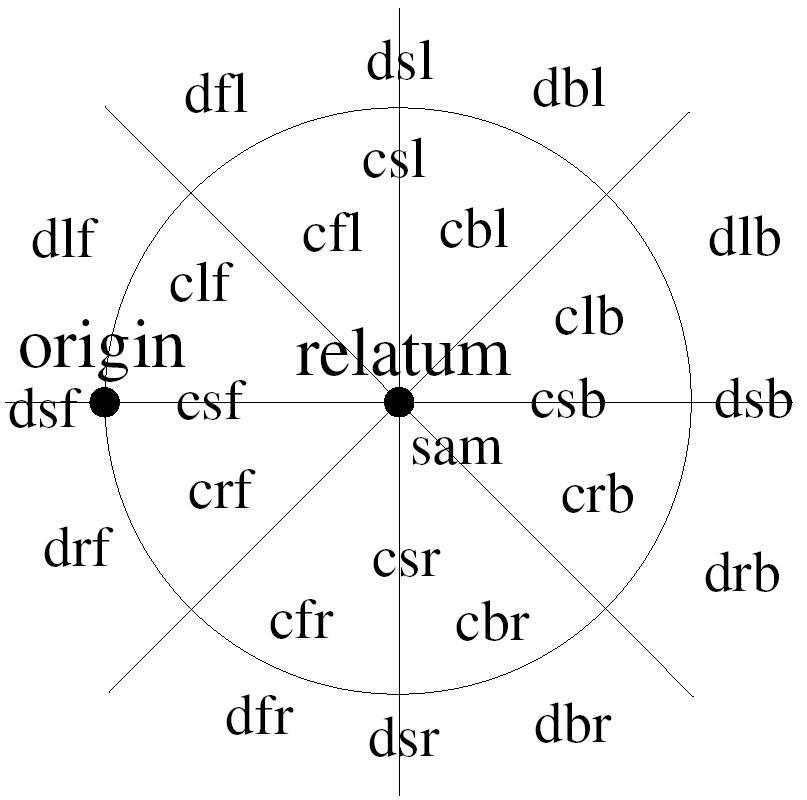}
\end{center}
	\end{captionabove}
\end{figure}

The TPCC calculus is derived from the cardinal direction calculus and provides finer differentiations than the cardinal direction calculus (see Section \ref{subsubsection:orientation}). Its 2D-plane is divided into eight slices by adding another cross which is rotated 45°. Additionally the distance from the relatum to the referent is compared to the distance from the relatum to the origin to provide a distinction between the two ranges. (See figure \ref{fig:TPCCnames})

The letters f, b, l, r, s, d, c in figure \ref{fig:TPCCnames} stand for front, back, left, right, straight, distant, close. The TPCC has 8 different orientations and 4 precise orientations. With 2 distances and three special cases there are 27 possible configurations ( (8 + 4) $\cdot$ 2 + 3 = 27).

For the mathematical definition of the \tpcc please see the appendix \ref{cap:TPCCdef} on page \pageref{cap:TPCCdef}.

Sets of TPCC definitions can be used as described in 
. The notation for set configurations is to write the relations of the set in parentheses, separated by commas - e.g. 
$A, B \mbox{ clf } C \vee A, B, \mbox{ cfl } C$ would be described by $A, B \mbox{ (clf,cfl) } C$.
\begin{figure}[ht]
	\begin{captionabove}{Iconic representation for TPCC-relations}
  	\label{fig:TPCCcross}
\begin{center}
		\includegraphics[width=5cm]{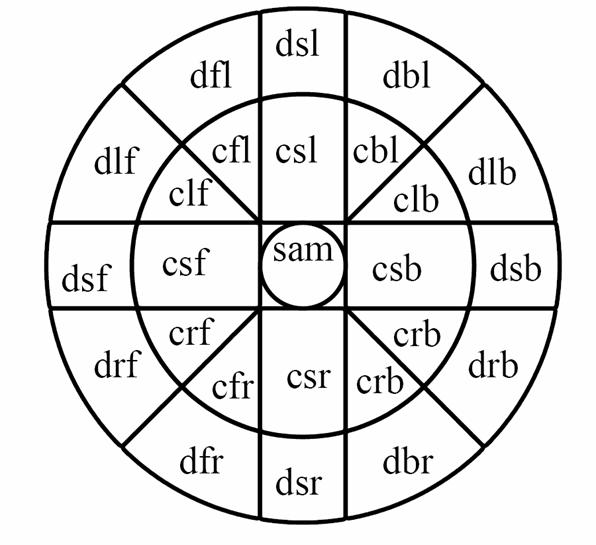}
\end{center}
	\end{captionabove}
\end{figure}

\subsection{Reasoning with TPCC}

The unary and binary operations introduced in sections \ref{sec:spatialReasoning} and  \ref{sec:qualitativespatialreasoning} are used with TPCC as well. In order to keep the transformation and composition tables small Moratz introduced an iconic representation for the TPCC-relations seen on table \ref{fig:TPCCcross}. Segments corresponding to a relation are illustrated as filled segments, sets of base relations have several segments filled. This representation is easier to translate into its semantic content compared with a representation that uses the textual relations symbols. 

\begin{figure}[ht]
	\begin{captionabove}{The iconic representation of the unary TPCC operations}
  	\label{fig:TPCCunaryops}
\begin{center}
		\includegraphics[width=8cm]{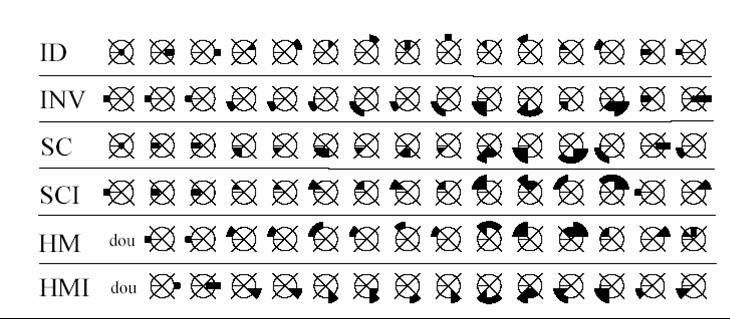}
\end{center}
	\end{captionabove}
\end{figure}

The transformation table \ref{fig:TPCCunaryops} shows that results of transformations may constitute subsets of the base relations. 

{\Large Composition}

TPCC is not closed under strong composition. Therefore a weak composition operation was introduced with the specific relation such that:

 $\forall A,B,D: A,B\left(r_1 \oplus r_2\right) D \leftarrow \exists C: A,B\left(r_1\right)C \wedge B,C\left(r_2\right)D$

The table for the weak composition can be found in \cite{moratzTPCC}. More than 700 results of the weak composition are represented in this table. 

{\Large Constraint-based Reasoning}

\csp~are solved as described in section \ref{sec:CSP}. To express the calculus in terms of relation algebras is one prerequisite for using this standard constraint algorithm. But the \tpcc~is not closed under transformations and under compositions so this algorithm doesn't achieve path-consistency. Nevertheless, it is possible to perform simple path-based inferences. This is done when the last two relations of a path are composed and the reference system is incrementally moved toward the beginning of the path in a form of a backward chaining. This can be used to detect cyclic paths. 
Moratz showed that reasoning with TPCC is in PSPACE.

%
%
%
%
%
\section{\doi}
\label{sec:doi}
The \doi (DOI) by Reinhard Moratz \cite{cosy_Moratz_03_Spatial} proposes an approach to model the typically imprecise sensor data about orientations and distances. This approach propagates orientation and distance intervals to produce global knowledge.

\subsection{DOI Introduction}

The \doi (DOI) was designed as a calculus for mobile robot indoor exploration. Therefore it reflects the imprecise sensor data robots usually provide. Furthermore the data delivered by the sensors may be not only imprecise but also incomplete. This poses serious problems for the integration of local spatial knowledge into survey knowledge which is, for example, required for robot explorations in unknown environments. Navigation tasks require calculi that handle orientation and distance information - pure topological information is not sufficient. \cite{roefer1999}

The DOI uses a relative reference system which utilizes continuous interval borders for modeling imprecision. Therefore it is not a qualitative calculus but it can be seen as an extension to the qualitative approaches. Benefits of qualitative approaches are combined with metric measurements by using DOI. Qualitative calculi can represent imprecise spatial knowledge while metric representations are good at distinguishing different spatial entities. The DOI propagation can be combined with a representation based on the generalized Voronoi graph of a robots free space developed by Wallgrün. The DOIs are used here  to specify the relative positions of the vertices in the Voronoi graph. 

\subsection{The DOI definition}

The \doi~is based on continuous distance orientation-intervals. There is again, on a 2D-plane,  a point-like object. The DOI uses this point together with a reference direction as anchor and it has four additional parameters: $r^{min}, r^{max}, \phi^{min}$ and $\phi^{max}$. 

A DOI d is a set of polar vectors ($r^i,\phi^j$) with:\\\\
$d = \left\{\left(r^i,\phi^j\right)|r^{min} \leq r^i \leq r^{max} \wedge \phi^{min} \leq \phi^j \leq \phi^{max}\right\} $\\

\begin{figure}[ht]
	\begin{captionabove}{A DOI and its parameters}
	  \label{fig:DOIdef}
\begin{center}
		\includegraphics[width=6cm]{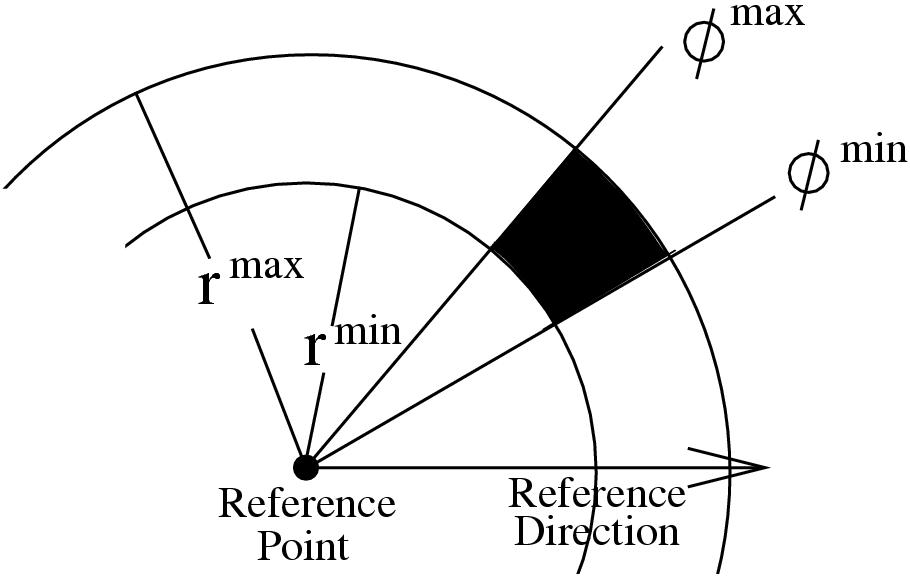}
\end{center}
	\end{captionabove}
\end{figure}

There is a special case which represents the spatial arrangement where the goal location can be the same as the reference point. This case is represented by the following values: $\phi^{max} = \pi$, $\phi^{min} = -\pi$ and $r^{min} = 0$.

In all other cases $\phi^{max}-\phi^{min} \leq \pi$ holds and for convenience it is assumed that $-2\pi \leq \phi^{min} \leq \pi$ and $-\pi \leq \phi^{max} \leq \pi$

\subsection{DOI composition}
\begin{figure}[ht]
	\begin{captionabove}{Composition of adjacent path segments}
		\label{fig:DOIcomp}
\begin{center}
  	\includegraphics[width=6cm]{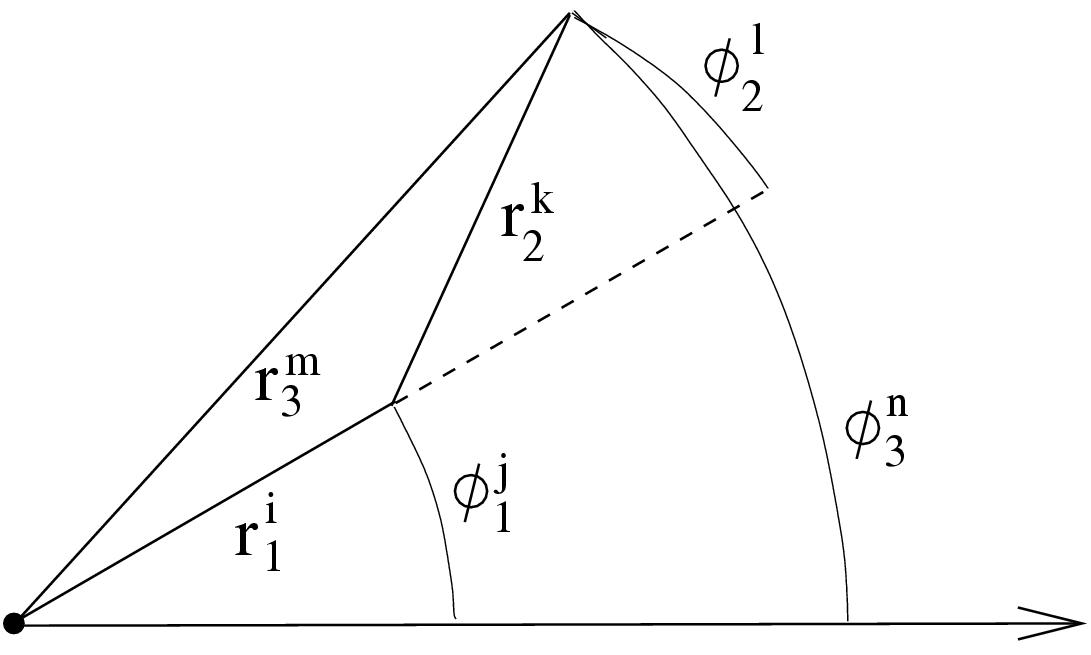}
\end{center}
	\end{captionabove}
\end{figure}

The composition between two DOIs is the basic step for propagation along paths (see figure \ref{fig:DOIcomp}). 

Two DOIs ($d_1 and d_2$) can be compositioned into a third ($d_3$).
\begin{eqnarray*}
d_1 & = & \left(r_1^{min},r_1^{max},\phi_1^{min},\phi_1^{max}\right)\\
d_2 & = & \left(r_2^{min},r_2^{max},\phi_2^{min},\phi_2^{max}\right)\\
d_3 & = & \left(r_3^{min},r_3^{max},\phi_3^{min},\phi_3^{max}\right)
\end{eqnarray*}
It holds:
\begin{eqnarray*}
d_3 & = & \Big( {min\atop r_{1}^i  ,r_2^k, \phi_1^j, \phi_2^l} 
r_{\sum}\left(r_1^i,r_2^k,\phi_1^j,\phi_2^l\right), {max\atop r_1^i,r_2^k,\phi_1^j,\phi_2^l}  r_{\sum}\left(r_1^i,r_2^k,\phi_1^j,\phi_2^l\right)\\
 & &{min\atop r_1^i,r_2^k,\phi_1^j,\phi_2^l} \phi_{\sum}\left(r_1^i,r_2^k,\phi_1^j,\phi_2^l\right),{max\atop r_1^i,r_2^k,\phi_1^j,\phi_2^l} \phi_{\sum}\left(r_1^i,r_2^k,\phi_1^j,\phi_2^l\right)\Big) 
\end{eqnarray*}

with: $ r_1^{min} \leq r_1^i \leq r_1^{max}, \phi_1^{min} \leq \phi_1^j \leq \phi_1^{max},
 r_2^{min} \leq r_2^i \leq r_2^{max}, \phi_2^{min} \leq \phi_2^j \leq \phi_2^{max}$\\

The functions $r_{\sum}$ and $\phi_{\sum}$ are defined this way:\\

$r_{\sum}\left(r_1^i,r_2^k,\phi_1^j,\phi_2^l\right) = \sqrt{
\left(r_1^i \sin \phi_1^j + r_2^k \sin \left(\phi_1^j + \phi_2^l\right)\right)^2 + 
\left(r_1^i \cos \phi_1^j + r_2^k \cos \left(\phi_1^j + \phi_2^l\right)\right)^2 }
$\\ 

and\\
$\phi_{\sum}\left(r_1^i,r_2^k,\phi_1^j,\phi_2^l\right) =
\tan^{-1} \frac{r_1^i \sin\phi_1^j + r_2^k \sin\left(\phi_1^j + \phi_2^l\right)}
			        {r_1^i \cos\phi_1^j + r_2^k \cos\left(\phi_1^j + \phi_2^l\right)}
$

The values for the minimum and maximum of $d_3$ thus the cases for which $r_3^m$ and $\phi_3^n$ have their minimum or maximum and can be listed by geometric analysis. These formulas are given in appendix \ref{cap:DOIcompform}.

\begin{figure}[ht]
	\begin{captionabove}{Resulting DOI}
  	\label{fig:DOIcomp2}
\begin{center}
		\includegraphics[width=6cm]{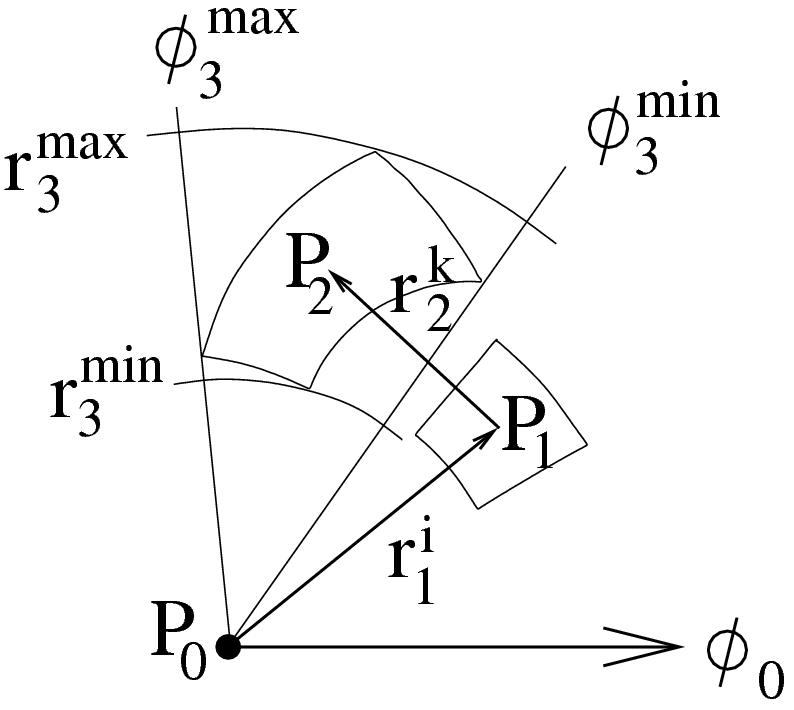}
\end{center}
	\end{captionabove}
\end{figure}

The composition is only an approximation in form of an upper bound of the area consisting of the vectors $(r_3^m, \phi_3^n)$ which can be directly composed by vectors $(r_1^i, \phi_1^j)$ and $(r_2^k, \phi_2^l)$ from $d_1$ and $d_2$ respectively. A typical spatial layout of these areas is shown in figure \ref{fig:DOIcomp2}.

A DOI can be viewed as a set of difference vectors between two points expressed in polar coordinates relative to the reference direction. A function $\delta$ maps two points and a reference direction to the corresponding difference vector. The function $\phi$ maps two points and a reference direction from the first point to the second. The relation between points can now be expressed in imprecise relative position and the respective DOIs:\\

$\Delta\left(\phi_0,P_0,P_1\right) \in \mbox{d}_1 \wedge \Delta\left(\phi\left(P_0,P_1\right),P_1,P_2\right) \in \mbox{d}_2 \Rightarrow \Delta\left(\phi_0,P_0,P_2\right) \in \mbox{d}_1 \diamond \mbox{d}_2$

Now the DOI composition can be used to propagate local, relative spatial knowledge along a path. There is an anchor point $P_0$, a reference direction $\phi_0$ and a sequence of points which determine the path segments $P_1$,$P_2$, \ldots $P_i$, \ldots $P_n$. Each Point $P_i$ on the path has an associated DOI $\mbox{d}_i$. A stepwise composition recursively beginning with the end of the path yields the relative position of the end point with respect to the anchor point $P_0$ and the reference direction $\phi_0$:\\
$\Delta\left(\phi_0,P_0,P_n\right) \in \mbox{d}_1 \diamond \left(\mbox{d}_2 \left(\ldots\left(\mbox{d}_{n-1} \diamond \mbox{d}_n\right)\ldots\right)\right)$

\cite{cosy_Moratz_03_Propagation}
\cite{cosy_Moratz_03_Spatial}

%
%
%
%
%
\section{Granular Point Position Calculus}
\label{sec:gppc}

In \cite{moratzGPPC} Reinhard Moratz developed a calculus called Granular Point Position Calculus (\textsc{Gppc}). Two points define a relative reference system in two dimensional space. The calculus is partitioning the space in several orientations and distances.\par  

The special cases for the origin $A=(x_A, y_A)$, the relatum $B=(x_B,y_B)$ and the referent $C=(x_C,y_C)$ are:\par
$A, B$ dou $C := x_A = x_B \wedge y_A = y_B \wedge (x_C \neq x_A \vee y_C \neq y_A)$\\
$A, B$ tri $C := x_A = x_B = x_C \wedge y_A = y_B = y_C$\par

The relative radius for the other cases is defined as:\par
$r_{A,B,C} := \frac{\sqrt{(x_C-x_B)^2+(y_C-y_B)^2}}{\sqrt{(x_B-x_A)^2+(y_B-y_A)^2}}$\par
$A, B$ sam $C := r_{A,B,C} = 0$\par
For $A \neq B \neq C$ the relative angle is defined as:\par
$\phi_{A,B,C} := tan^{-1}\frac{y_C-y_B}{x_C-x_B}-tan^{-1}\frac{y_B-y_A}{x_B-x_A}$\par

\begin{figure}[ht]
	\begin{captionabove}{Example configuration in \textsc{Gppc}}
  	\label{fig:gpcc}
\begin{center}
		\includegraphics[width=7cm]{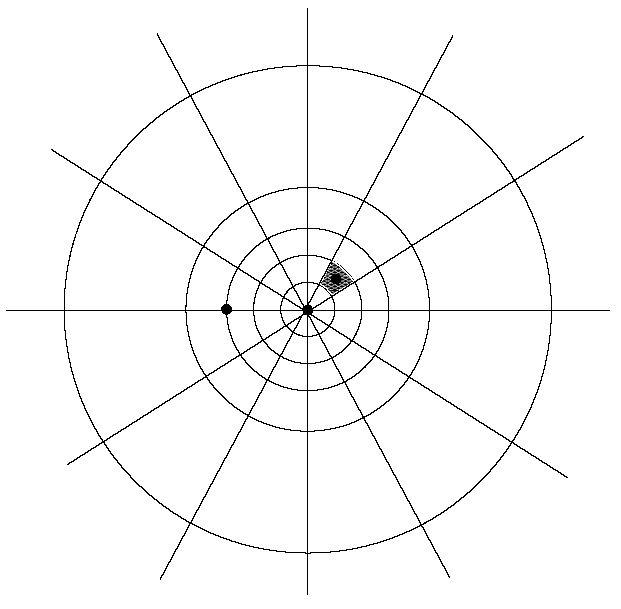}
\end{center}
	\end{captionabove}
\end{figure}

\textsc{Gppc} allows different levels of granularity and figure \ref{fig:gpcc} shows an example configuration of an \textsc{Gppc} with the granularity level m of three. A \textsc{Gppc}$_m$ has $(4m-1)(8m) + 3$ base relations which are defined as follows:\\
\begin{eqnarray*}
0 \leq j \leq 8m - 2 \wedge j \mbox{ mod } 2 = 0 & \rightarrow & \phi_{A,B,C} = \frac{j}{4m}\pi\\
1 \leq j \leq 8m - 1 \wedge j \mbox{ mod } 2 = 1 & \rightarrow & \frac{j-1}{4m}\pi < \phi_{A,B,C} < \frac{j+1}{4m}\pi\\
1 \leq i \leq 2m - 1 \wedge i \mbox{ mod } 2 = 1 & \rightarrow & \frac{i-1}{2m} < r_{A,B,C} < \frac{i+1}{2m}\\
2 \leq i \leq 2m \wedge i \mbox{ mod } 2 = 0 & \rightarrow & r_{A,B,C} = \frac{i}{2m}\\
2m + 1 \leq i \leq 4m - 3 \wedge i \mbox{ mod } 2 = 1 & \rightarrow & \frac{m}{2m-\frac{i-1}{2}} < r_{A,B,C} < \frac{m}{2m-\frac{i+1}{2}}\\
2m + 2 \leq i \leq 4m - 2 \wedge i \mbox{ mod } 2 = 0 & \rightarrow & r_{A,B,C} = \frac{m}{2m-\frac{i}{2}}\\
i = 4m - 1 & \rightarrow & m < r_{A,B,C}\\
\end{eqnarray*}
The DOI calculus presented in the section above is used to calculate the composition table for the \textsc{Gppc}. This is possible because the flat segments and their borders are summarized obtaining a quasi-partition. \cite{moratzGPPC}

%
%
%
%
%

%% file: chapters/FSPPapproach.tex
%
%
%
%
\chapter{The \fs~ Approach}
\label{sec:FSPP approach}







The calculus that is being developed in this thesis designed for usage in indoor robotics, but it can be used for other applications that reason about positional information, as well. 

For current robotic navigation and communication systems it is sufficient to work with two dimensional space. Reasons for that are, that the movement most often happens on the ground, that the sensors as well as the maps are two dimensional and also because humans most often communicate this way if there is not a good reason to do otherwise. Of course two dimensional reasoning is much easier than three dimensional reasoning, while it is simply not possible to do two dimensional navigation with one dimensional calculi. \\
It is also assumed that the space is isometric and homogeneous. Isometric means, that the cost of moving is the same in all directions while a homogeneous space is one that has the same properties at all locations. These constraints are made because otherwise it would be really difficult to develop an adequate representation.\par
The most important spatial aspects for distinguishing objects are shape, topology and position. Representing the shape of an object is independent from its position and vice versa if an absolute or relative reference system is used. Therefore it does not influence the calculus developed here and it is thus disregarded for it. It is further assumed that all considered spatial entities are disjoint. Reasons for that are, that sensors anyways have problems to distinguish objects that meet each other, that most objects actually are disjoint from each other and if they shouldn't be disjoint it is most often not important to recognize that fact. Therefore the most interesting spatial aspect of space for robotics is the position, consisting of orientation information and distance information \cite{musto}.\par


There are already some calculi representing orientation, distance or position information, as it has been shown in chapter \ref{sec:StateOfTheArt}. Some of them already have been used quite successfully in robotic context \cite{Dialog}. Nevertheless the new approach for representing and reasoning about positional information presented here should be advantageous over these calculi.\par
The Dipole calculus premises extended objects with an intrinsic orientation which are quite difficult to recognize for sensors. As discussed in section \ref{sec:qsr} it is more convenient to use points for representing orientation and distance.\\
The \tpcc~has already been used in some applications. For the given tasks it was rather successful but it became apparent that is was too coarse in many situations.  The 8 orientations and two distances generate 16 flat acceptance areas in TPCC - too less in scenarios with many object or greater distances.\\
The \doi~has a quite accurate representation for single objects. It is however not a qualitative approach. It doesn't support disjunctions of base relations and has thus difficulties in representing different possible positions - those have to be fitted into the characteristicly shaped DOI acceptance area which may mean a loss of precision.\\
The Granular Point Position Calculus is advantageous over the three latter approaches. It is a qualitative approach that represents point position with an arbitrary level of granularity in both distance and orientation. Although it is quite similar to the approach developed in this thesis, it differs in some important points from the \textsc{Fspp}. A comparison of these two calculi will be made in section \ref{sec:comparison}.\par

The name of the new calculus is "Fine-grained Qualitative Spatial Reasoning about Point Positions" (\fs). It already highlights the major properties of the representation.\par

\section{\fs~Representation}

\fs~has an arbitrary level of granularity. Depending on the scenario a very fine-grained representation could be chosen or a quite coarse one which would be less memory and time consuming. Very little changes need to be done in order to change the granularity.

The \fs~is defined over ternary points: the origin, the relatum and the referent (see section \ref{subsubsection:orientation}). The three points are needed for the qualitative orientation, because it will use a relative frame of reference. The qualitative distance will, however, make not use of the origin as you will see in the next section.

\subsection{\fs~Distance}
\label{sec:fsppDist}
Hernández distance definitions from \cite{hernandez1995} and \cite{hernandez1997}, which already have been summarized in section \ref{sec:qsrDistance}, will be used for \fs, too. 

\begin{figure}[ht]
  \begin{captionabove}{Distance from relatum $\Delta_x$ and distance range $\delta_x$}
	  \label{fig:distanceRange}
\begin{center}
		\includegraphics[height=5.5cm]{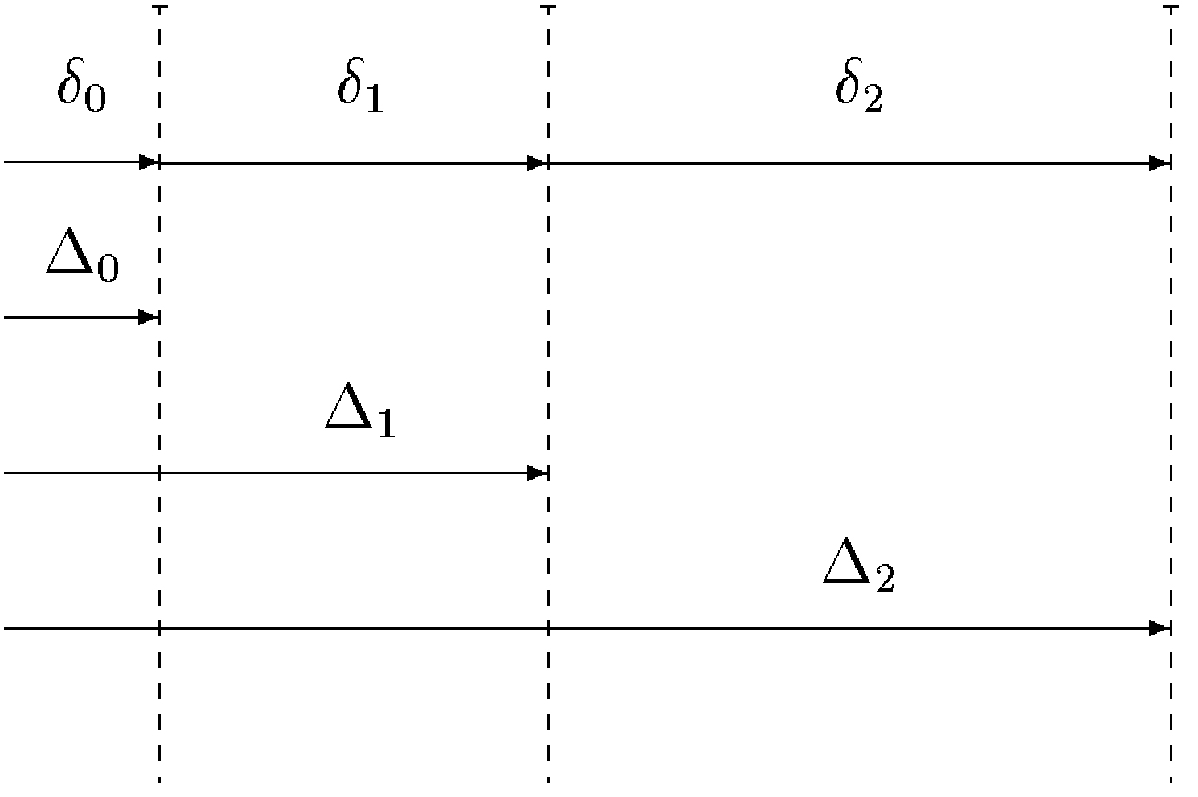}
\end{center}
	\end{captionabove}
\end{figure}

The distance system $D$ is defined over a totally ordered set of distance relations $Q$, an acceptance function $A$ and an algebraic structure $\Im$: $D=(Q,A,\Im)$ (see section \ref{sec:qsrDistance}). $Q = \{q_0,q_1,q_2,...,q_n\}$, given a level of granularity with $n+1$ distance distinctions. The width of an acceptance area corresponding to a distance symbol $q_i$ is denoted with $\delta_i$ whereas $\Delta_i$ denotes the maximum distance that a relatum can have to the referent for still falling into the acceptance are of the distance symbol $q_i$ (see figure \ref{fig:distanceRange}).

\fs~uses an absolute distance representation. The area around the relatum is partitioned into a number of circular acceptance areas using a global scale. In the indoor robotics scenario the action radius is quite low compared to the distance of objects within the robots sensor input. It is thus acceptable to use a monotone distance system. 

The absolute representation was chosen because the quantitative data input (from sensors or maps) allows an easy abstraction to an absolute distance representation (opposed to orientation). It is desirable to keep those absolute distance information, because it is very difficult to find out an metric distance for the robot using only relative distance relations. Absolute distances also yield in better composition results and easier usage in maps and other high level applications. A drawback is the more difficult human-robot communication.

For the distance part of \fs~the origin is not needed. The distance is solely defined over the relatum and the referent. That leads to the fact that all unary operations except \textsc{Id} and \textsc{Sc} (see table \ref{tab:transformations}) loose all distance information - the resulting \fs s only have orientation information, all distances are possible for any possible orientation. But that is no disadvantage. During the constrain propagation process (see section \ref{sec:CSP}) the distance information will be restored if it has been available in the initial set of constraints (relations). 

In the absolute distance system for length $L$ of the first interval $\delta_0$ has to be specified. It's value strongly depends on the number of distance distinctions $m$ as well as on the distance system chosen. The following values seem to be quite plausible and are thus used later on:
\begin{eqnarray*}
L & = & 10cm\\
m & = & 24\\
\delta_{i+1} & = & \delta_i \cdot 1.25
\end{eqnarray*}

The last distance interval $q_{24}$ has a size of $\delta_{24} = 26.5m$ ($1.25^{25} \cdot 10cm$) and the maximum distance would be at $\Delta_{24} = 132m$. The \fs~should consider distances greater than the maximum distance to fall into the last distance interval. The size of $\delta_{24}$ and $\Delta_{24}$ is thus infinite. Besides that exception all distance intervals $\delta_i$ bigger than ten ($i > 10$) than have about 20\% of the size of the total distance $\Delta_i$. This seems to be a good value - but it isn't suggested by cognitive articles known to the author.

\subsection{\fs~Orientation}
\label{sec:fsppOri}
\fs~uses a relative frame of reference for orientation. The advantages of relative positions are, that no intrinsic characteristics of objects have to be distinguished and that no global coordinate system has to be used which is often not available in indoor scenarios. 

The space around the relatum is partitioned according to a number of ordered orientation distinctions $R = \{o_0,o_1,o_2,...,o_m\}$. The \textit{acceptance function} defines acceptance areas around the relatum which correspond to the orientation relations ($o_j$).

Depended on the granularity chosen, different numbers of orientation relations are possible. The number of orientations $m$ is always even. This way the origin is always located on the border between the acceptance areas of the orientation relations $o_{\frac{m}{2}-1}$ and $o_{\frac{m}{2}}$. These special cases where objects are located directly on the border between two acceptance areas are very unlikely to happen in real life scenarios. It is thus acceptable to define, that in such cases all bordering acceptance areas are to be considered as possible locations for the referent. This is called quasi-partition. 

The acceptance function for an angle $phi$ are defined by the following formula:

\begin{center}
$ \phi $ in $ o_j \rightarrow \frac{\pi}{m}j \leq \phi \leq \frac{\pi}{m}(j+1)$
\end{center}

\subsection{The Position}
\label{sec:fsppPos}
Now the distance and the orientation are put together. This is done by defining the Cartesian product of the set of distance relations $Q$ and the set of orientation relations $R$: $S = Q \times R$. Thus \fs~has a set $S$ of $m \cdot n$ atomic relations $f_{i,j}$:

$S = \left(
\begin{array}{*{4}{c}} 
f_{0,0} & f_{0,1} & \cdots & f_{0,j} \\
f_{1,0} & f_{1,1} & \cdots & f_{1,j} \\
\vdots & \vdots & \ddots & \vdots \\
f_{i,0} & f_{i,1} & \cdots & f_{i,j} \\
\end{array}
\right)$

The \fs~is defined over three points: the origin $A$ with it's coordinates $A = (x_A, y_A)$, the relatum $B = (x_B, y_B)$ and the referent $C = (x_C, y_C)$. The following definitions are similar to those in the \textsc{Gppc} (see section \ref{sec:gppc}) because both calculi use similar acceptance areas.\par

First the three special cases that can occur in all ternary calculi are defined:\\

\begin{eqnarray*}
A, B \mbox{ dou } C & := & x_A = x_B \wedge y_A = y_B \wedge (x_C \neq x_A \vee y_C \neq y_A)\\
A, B \mbox{ tri } C & := & x_A = x_B = x_C \wedge y_A = y_B = y_C\\\\
&&\mbox{The relative radius for the other cases is defined as:}\\\\
r_{A,B,C} & := & \frac{\sqrt{(x_C-x_B)^2+(y_C-y_B)^2}}{\sqrt{(x_B-x_A)^2+(y_B-y_A)^2}}\\
A, B \mbox{ sam } C & := & r_{A,B,C} = 0\\\\
&&\mbox{For $A \neq B \neq C$ the relative angle is defined as:}\\\\
\phi_{A,B,C} & := & tan^{-1}\frac{y_C-y_B}{x_C-x_B}-tan^{-1}\frac{y_B-y_A}{x_B-x_A}
\end{eqnarray*}

The acceptance areas for the \fs~ representation are defined according to the distance and orientation relations above:\\

\begin{center}
$C \mbox{ in } f_{i,j} \mbox{  } \rightarrow \mbox{  } \Delta_{i-1} \leq r_{A,B,C} \leq \Delta_i \mbox{ }\wedge\mbox{ } \frac{\pi}{m}j \leq \phi_{A,B,C} \leq \frac{\pi}{m}(j+1)$
\end{center}
For the first and the last distances the following definitions are made:\\
$\Delta_{-1} = 0$\\
$\Delta_n =$ infinite\par

\begin{figure}[ht]  
	\begin{captionabove}{An example FSPP with 32 orientation distinctions and 13 distances}
  	\label{fig:FSPP1}
\begin{center}
		\includegraphics[height=7cm]{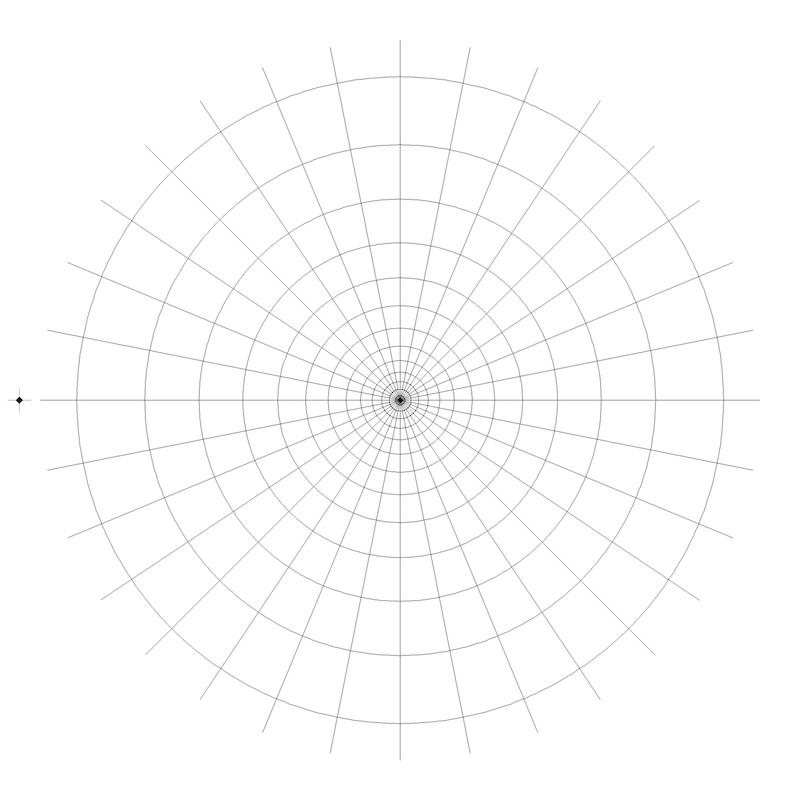}
\end{center}
	\end{captionabove}
\end{figure}

\begin{figure}[ht]  
	\begin{captionabove}{Disjunction of 22 \fs~base relations indicating possible locations of the referent}
  	\label{fig:FSPP2}
\begin{center}
		\includegraphics[height=5cm]{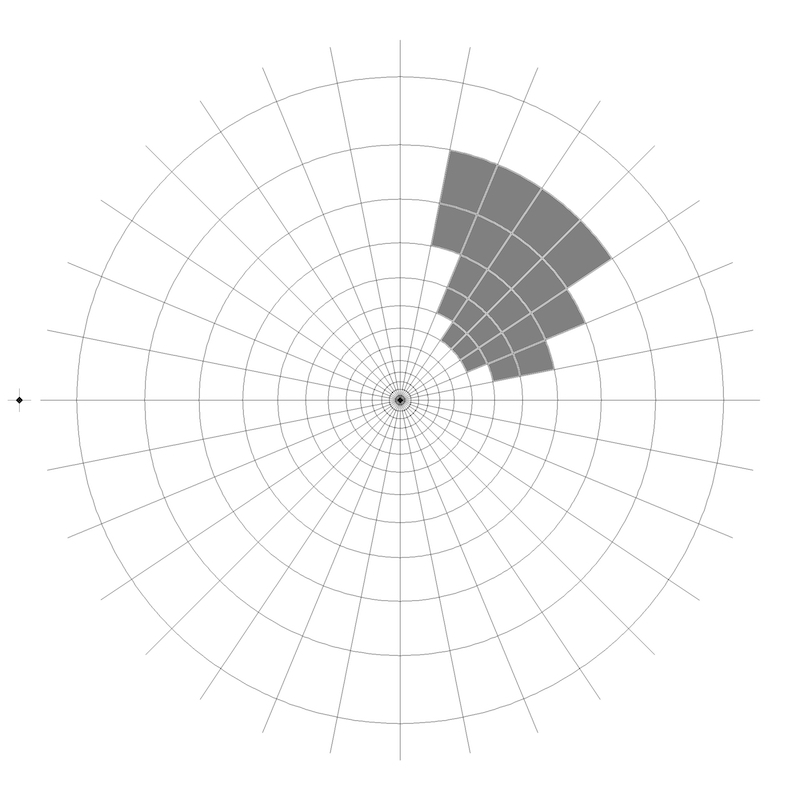}
\end{center}
	\end{captionabove}
\end{figure}

In figure \ref{fig:FSPP1} an \fs~with 416 (exempt the three special cases) base relations is shown. For display improvements only 13 distances are used rather than the suggested 24, but the distance system follows the  $\delta_{i+1} = \delta_i \cdot 1.25$ formula. 32 orientations are differentiated. The origin is indicated on the left (it's distance to relatum and referent is unknown), the relatum is in the middle while the referent is the object that is to be located. In figure \ref{fig:FSPP2} an example for possible locations of the referent are indicated gray. The referent object has equal probabilities of being in one of the 22 indicated base relations. The displayed \fs~relation thus consists of a disjunction of those 22 base relations. 

\section{\fs~Reasoning}
\label{sec:fsppReas}


The unary operations and the binary composition are needed for constraint based reasoning. But those operation are not trivial in the given representation. 

\subsection{Composition}
The \fs~was developed so that it can make use of the \doi~. The DOI provides easy access to composition results. The atomic relations of the \fs~are exactly shaped like DOIs so that it is no problem to use the DOI composition for the atomic relations. This is done on demand and not prior due to the huge composition table that would have to be created. The composition result of a general \fs~relation is then the union of the composition results of the atomic relations.\par

\subsection{Unary operations}
The unary operations are more difficult to develop. The problem is here, that the distance of the origin to the relatum and the referent is unknown which leads to quite ambiguous results. The unary operations are calculated as follows (the identity operation \textsc{Id} is trivial and thus exempted):\\

\subsubsection{Inversion \textsc{Inv}}
\begin{figure}[ht]  
	\begin{captionabove}{The temporary \fs~ for the \textsc{Inv} operation}
  	\label{fig:inv}
\begin{center}
		\includegraphics[height=5cm]{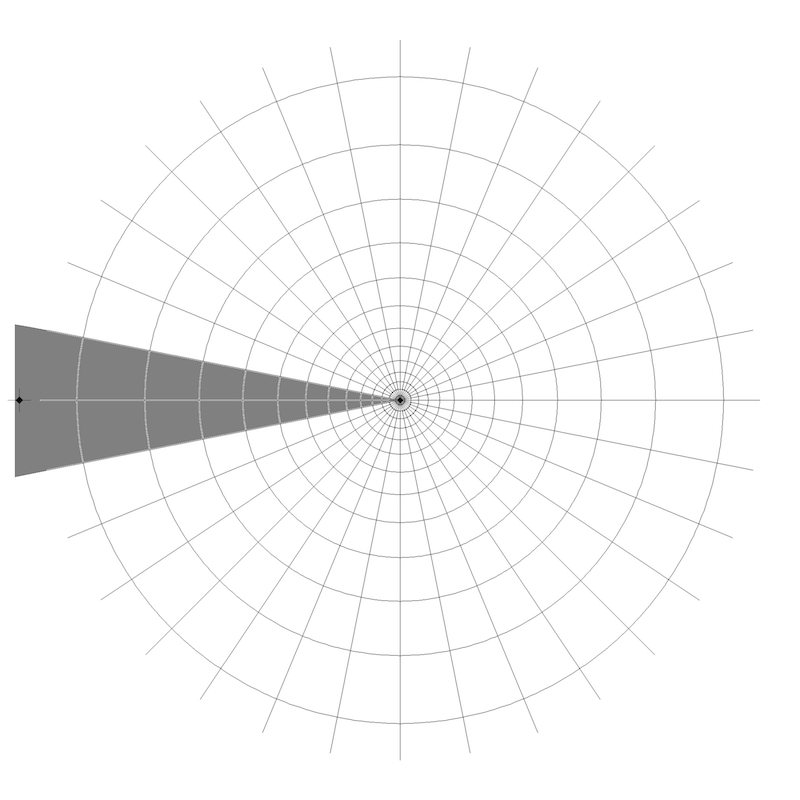}
\end{center}
	\end{captionabove}
\end{figure}

For the computation of the inversion the composition is used. First a temporary \fs~is created. It represents the possible location of the origin A to the relatum B - A is always at the border in the back and it has an unknown distance. The \fs~thus looks like in figure \ref{fig:inv}. Now the composition of this temporary \fs~with the original has the inversion as a result. 

\subsubsection{Short cut \textsc{Sc}}
The short cut reorders the \fs~(ABC) to (ACB). That means, that the distance information is preserved (since the distance BC is the same as CB). But the orientation is ambiguous after that operation because of the unknown distance of the origin A. The \textsc{Sc} is calculated as follows:\\

\begin{figure}[ht]  
	\begin{captionabove}{The short cut \textsc{Sc} operation}
  	\label{fig:sc}
\begin{center}
		\includegraphics[height=5cm]{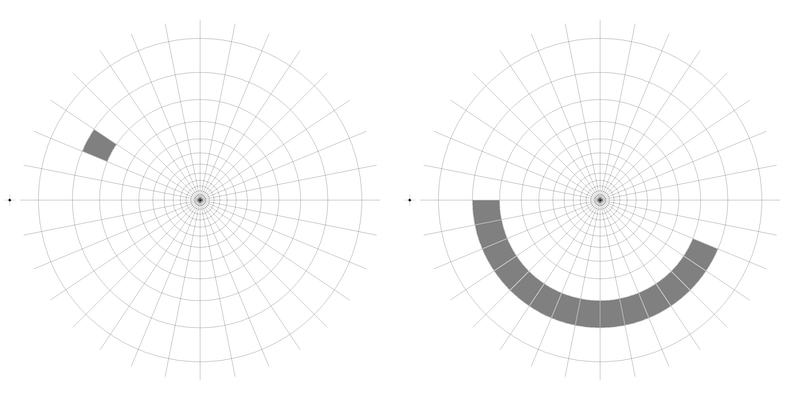}
\end{center}
	\end{captionabove}
\end{figure}

The result of \textsc{Sc}$(o_i)$ is for $i< \frac{m}{2}$: all orientations in the range from $\frac{m}{2}$ to $\frac{m}{2}+i$. For $i \geq \frac{m}{2}$: all orientations in the range from $\frac{m}{2}-1$ to $i-\frac{m}{2}$. Figure \ref{fig:sc} shows the original \fs~on the left and the result of the \textsc{Sc} on the right.

\subsubsection{\textsc{Sci}, \textsc{Hm} and \textsc{Hmi}}
The remaining unary operations can be calculated using those two above:\\

\begin{eqnarray*}
\mbox{\textsc{Sci}(ABC)} & \rightarrow & \mbox{\textsc{Inv}(\textsc{Sc}(ABC))}\\
\mbox{\textsc{Hm}(ABC)} & \rightarrow &  \mbox{\textsc{Sc}(\textsc{Inv}(ABC))}\\
\mbox{\textsc{Hmi}(ABC)} & \rightarrow &  \mbox{\textsc{Sc}(\textsc{Sci}(ABC))}\\
\end{eqnarray*}




\subsection{Conceptual neighborhood of the \fs~}
\begin{figure}[ht]  
	\begin{captionabove}{Conceptual neighborhood of an \fs}
  	\label{fig:conceptualN}
\begin{center}
		\includegraphics[height=5cm]{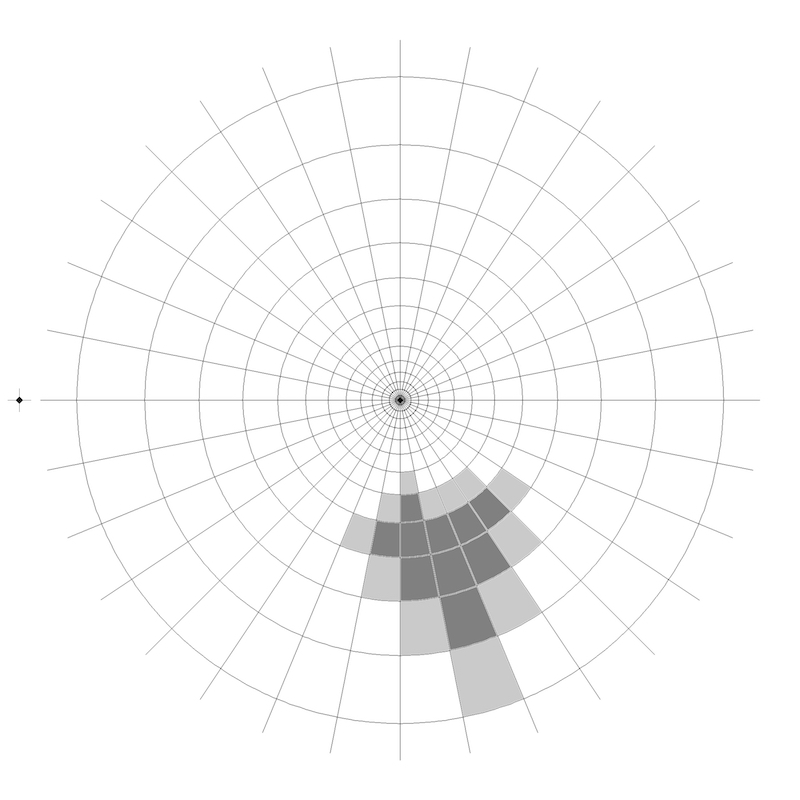}
\end{center}
	\end{captionabove}
\end{figure}

The definition of the conceptual neighborhood of the \fs~is quite straight forward. Movement is only possible between the acceptance areas that share an edge since quasi-partition is used. The conceptual neighbors of an \fs~ $f_{i,j}$ are thus $f_{i, (j+1)\mbox{mod}m}$, $f_{i, (j-1+m)\mbox{mod}m}$, $f_{i-1,j}$ and $f_{i+1}$. The first two conceptual neighbors are those with the same distance but neighboring orientations. It is necessary to use the modular operation since the orientation is circular - the "last" basic orientation relation is next to the "first" one. The latter two conceptual neighbors are those with the same orientation but with smaller and bigger distance. Those are only existent, of course, if $i>0$ respectively $r<n$. See figure \ref{fig:conceptualN} for an example. The dark gray atomic relations are the \fs~relation and the light gray relations are it's conceptual neighborhood. The geometric neighborhoods introduced in the next section have similarities with the conceptual neighborhood, but those concepts are defined over different basics and for different applications so that they are only partly comparable. 

\subsection{Improvement for the composition: Contour tracing}

Complexity issues are an important topic not only for \AI~ but also for qualitative spatial reasoning \cite{cosy:dylla:2004:TPCC_Complexity}. A quite easy way of reducing the number of computations needed will be presented in this section.\\

A problem in the composition approach described above is, that if the \fs~have more than a few basic relations, the number of DOIs that have to be computed is rapidly growing. Many of those calculations are unnecessary since they often lead to results (basic \fs~relations) that already have been found to be possible locations of the referent. A good way to avoid these computations is to only calculate the composition of those atomic relations that are bordering to atomic relations that are known to not contain the referent.\\

An efficient way of finding those relations will be shown in this section. The approach is called contour tracing, but it is also known as border or boundary following. In order to introduce the actual algorithm some definitions have to be made.

\subsubsection{Geometrical neighborhood}
The geometrical neighborhoods that are interesting for this thesis are defined on a two dimensional plane that is divided with square tessellation. The resulting squares are considered to have either a value of 0 or 1. Although the \fs~basic relations are no squares because two of their edges are round, they can still be used for these approaches. Since contour tracing is a computer graphics problem the squares are being referred as pixels which can have the color white for the value 0 and black for the value 1. That way it is also easier to understand the following terms and algorithm. 

\subsubsection{Moore Neighborhood}

\begin{figure}[ht]
	\begin{captionabove}{The Moore Neighborhood}
		\label{fig:MooreNeighborhood}
\begin{center}
		\includegraphics[width=3.5cm]{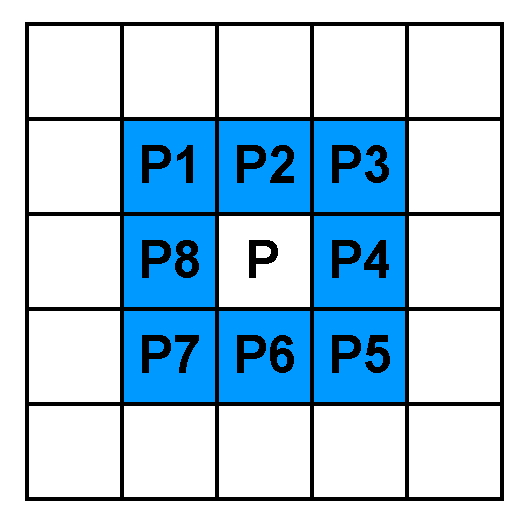}
\end{center}
	\end{captionabove}
\end{figure}

The Moore neighborhood of a pixel, $P$, is the set of 8 pixels which share a vertex or edge with that pixel. These
pixels are called $P1$, $P2$, $P3$, $P4$, $P5$, $P6$, $P7$ and $P8$ beginning with the top left one and circling clockwise around the pixel $P$ (see figure \ref{fig:MooreNeighborhood}). The Moore neighborhood (also known as the 8-neighbors or indirect neighbors) is an important concept that frequently arises in the literature.\par

\subsubsection{4-neighborhood}
The 4-neighborhood is a subset of the 8-neighborhood. It only consists of the pixels that share an edge with the pixel $P$. Those are the pixels $P2$, $P4$, $P6$ and $P8$ of the Moore Neighborhood.

\subsubsection{Border pixels}
Along the lines of neighborhoods the border pixels can be defined as 8-border pixels or 4-border pixels. A black pixel is an 8-border pixel if it shares a vertex or edge with at least one white pixel. Using the neighborhood definitions from above we can say that a black pixel is an 8-border pixel if at least one of the pixels of its Moore Neighborhood are white.\par
4-border pixels are thus black pixels which have at least one white pixel in their 4-neighborhood. Those pixels share an edge with one or more white pixels. 

\subsubsection{Connectivity}
A connected component is a set of black pixels, $B$, such that for every pair of pixels $p_i$ and $p_j$ in $B$, there exists a sequence of pixels  $p_i, ..., p_j$ such that:
\begin{enumerate}
\item all pixels in the sequence are in the set $B$ (they are black) and 
\item every 2 pixels that are adjacent in the sequence are neighbors. 
\end{enumerate}
A component is 4-connected if 4-neighborhood used in the second condition and it is 8-connected if the Moore Neighborhood is used.  

\subsubsection{Pavlidis contour tracing algorithm}

In 1982 Theo Pavlidis introduced his contour tracing algorithm \cite{Pavlidis}. With this algorithm it is possible to extract 4-connected borders as well as 8-connected ones. It is important to keep track of the direction in which you entered the pixel which is currently active. 

\begin{figure}[ht]
	\begin{captionabove}{Pavlidis: check $P1$}
		\label{fig:pla1}
\begin{center}
		\includegraphics[width=3.5cm]{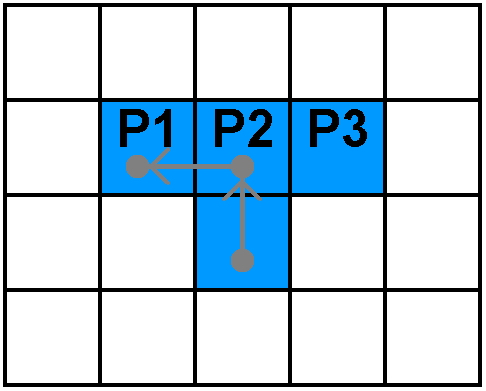}
\end{center}
	\end{captionabove}
\end{figure}

\begin{figure}[ht]
	\begin{captionabove}{Pavlidis: check $P2$ (left) and check $P3$ (right)}
		\label{fig:pla2}
\begin{center}
		\includegraphics[width=3.5cm]{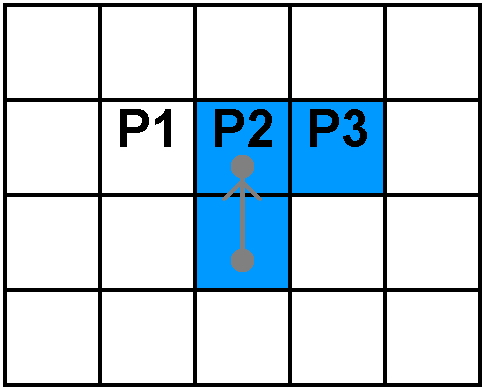}\hspace{1cm}
		\includegraphics[width=3.5cm]{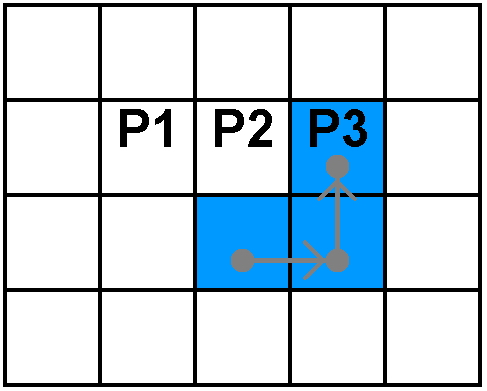}
\end{center}
	\end{captionabove}
\end{figure}

The first task that has do be done using this algorithm is to find a black start pixel. The only restriction to this pixel is, that its left neighbor ($P8$ in the Moor Neighborhood model) is white. From now on the only important pixels are the ones in front of the current pixel (left, ahead and right front: $P1$, $P2$ and $P3$). Four different cases have to be considered now:
\begin{enumerate}
\item If $P1$ is black it is declared as our new current pixel and the current direction is changed by turning left (-90°). $P1$ is also added to the set of border pixels $B$. (see figure \ref{fig:pla1})
\item If the first failed ($P1$ is white) and $P2$ black the new current pixel is $P2$ which is also added to the set of border pixels $B$. The current direction is not being changed. (see figure \ref{fig:pla2})
\item If the first two cases failed ($P1$ and $P2$ white) but $P3$ is black the new current pixel is $P3$. $P3$ is added to the set of border pixels $B$ and the direction is not being changed. (see figure \ref{fig:pla2}).
\item If first three cases failed (all three pixels $P1$, $P2$ and $P3$ are white) the current direction is changed by turning right (90°). 
\end{enumerate}
These above computations are repeated till one of the following two exit conditions is met:
\begin{enumerate}
\item The algorithm will terminate after turning right three times \textbf{on the same pixel} or,
\item after reaching the start pixel.
\end{enumerate}

The above algorithm returns a border that is 8-connected. If a 4-connected border is needed the first and the third case have to be altered. For the first case pixel $P2$ has to be black, too, or the first case fails. If $P2$ is black it also has to be added to $B$ after proceeding as stated above. In the third case $P4$ has to be black, too, or this case fails. If it is black is has to be added to $B$, too, and the algorithm proceeds as above. 

\cite{Pavlidis}

\subsubsection{Using Pavlidis algorithm in \fs}

\begin{figure}[ht]  
	\begin{captionabove}{8-connected border of an \fs}
  	\label{fig:border}
\begin{center}
		\includegraphics[height=5cm]{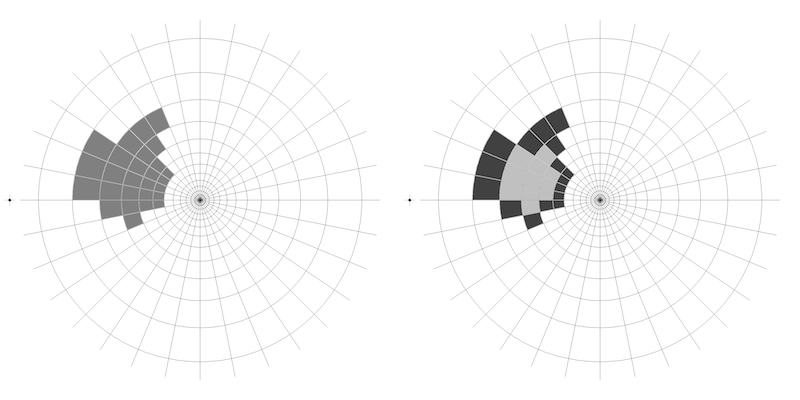}
\end{center}
	\end{captionabove}
\end{figure}

For \fs~the 8-connected border is sufficient because it corresponds to the conceptual neighborhood of the calculus. Figure \ref{fig:border} shows a random \fs~on the left. On the right the border atomic relations of this random \fs~are indicated dark while the inner atomic relations are light gray.\\

The result of a composition that used at least one \fs~with border segments is likely to not be solid but to have holes in it's representation that need to be filled (i.e. atomic relations that are needed to be activated). For this task Pavlidis algorithm is used again. This time not only the border relations are stored but also those outside atomic relations that are neighbors of the border segments and that do not contain the referent. Those outside segments then enclose the the actual \fs~relation. Now all atomic relations, that are neighbors of relations that are already activated, are activated, too, if they are not one of those enclosing segments calculated above. This way a fast filling algorithm for the \fs~is implemented.\\

During the constraint propagation it is only needed to fill the \fs~when doing intersections. For all other operations it is sufficient to use the border relations. It thus might be a good idea not to use the contour tracing algorithm after every operation but maybe only after every second or third. At the end of the inference process the resulting \fs{}s should be filled.

\section{\fs~Impelementation}
\label{sec:impl}
The \fs~is implemented in C. It uses a \doi~implementation which also has been written in C by the author. The representation of the basic relations is done by setting bits in an byte array. The naming conventions in the implementation are different to those in this theses, mainly because the implementation started before this document reached the actual \fs~chapter. The number of orientations $m$ is defined as "ROT\_SLICES" (for rotation) and the number of distances $n$ is defined as "TRANS\_SLICES" (for translation). \fs{}s are named "Pieces". The various input and output functions come pairwise - the ones with an A at the end request the orientation and the distance index of the atomic relation whereas the ones with an B at the end request the bit-number of the atomic relation. The \verb|BYTE_NUM(x)		(x/8)| and \verb|BIT_NUM(x)		(x%8))| directives are used to access the actual bit in the array. Some short exemplary functions are shown to demponstrate the \fs~implementation.

The procedures to set a specific atomic relation to an value (either 0 or 1) look like this:
{\small
\begin{verbatim}
void setA(Pieces *pcs, int rot, int trans, int value){
    setB(pcs,rot+trans*ROT_SLICES,value);
}

void setB(Pieces *pcs, int which, int value){
    if(value==0){
        pcs->pPcs[BYTE_NUM(which)] = 
            ~(1 << BIT_NUM(which)) & pcs->pPcs[BYTE_NUM(which)];
    }else{
        pcs->pPcs[BYTE_NUM(which)] = 
            1 << BIT_NUM(which) | pcs->pPcs[BYTE_NUM(which)];
    }
}
\end{verbatim}}
The get functions:
{\small\begin{verbatim}
int getA(Pieces *pcs, int rot, int trans){
    return getB(pcs, rot+trans*ROT_SLICES);
}

int getB(Pieces *pcs, int which){
    return (((1 << BIT_NUM(which)) & pcs->pPcs[BYTE_NUM(which)]) > 0);
}
\end{verbatim}}
The union of two FSPPs:
{\small\begin{verbatim}
void pieceORpiece(Pieces *rtn, Pieces *pcs1, Pieces *pcs2){
    int i;

    // setting the last unused bits in the last byte to 0
    pcs1->pPcs[BYTES_USED-1] &= ~(255<<BIT_NUM(PIECES));
    pcs2->pPcs[BYTES_USED-1] &= ~(255<<BIT_NUM(PIECES));

    for(i=0; i<BYTES_USED; i++){
        rtn->pPcs[i] = (pcs1->pPcs[i] | pcs2->pPcs[i]);
    }
}
\end{verbatim}}
The composition using DOI:
{\small\begin{verbatim}
void composition(Pieces *rtn, Pieces *pcs1, Pieces *pcs2){
    DoiTyp doi1, doi2, comp;
    Pieces tmp;
    Searchy search1,search2;
    int x,y;

    tmp.pPcs = getPiecesArray();

    search1.pPiece = pcs1;
    search1.number = -1;    // initializing the serach 
    search2.pPiece = pcs2;
    // searching through the first FSPP for atomic relations
    while((x = getNextTrue(&search1))>= 0){
        search2.number = -1; // initializing the serach 
        // searching through the second FSPP for atomic relations
        while((y = getNextTrue(&search2))>= 0){
            singleDoiB(&doi1,x); // generating a DOI with 
                                 // the atomic relation x
            singleDoiB(&doi2,y); // generating a DOI with
                                 // the atomic relation y
            // the composition
            composition_exact(&doi1,&doi2,&comp,""); 
            clearPiece(&tmp);
            // now a FPSS is being generated out of 
            // the DOI composition result
            doiToAnder(&tmp,&comp);
            // the union of this atomic composition result
            // with the former composition results
            pieceORpiece(rtn,rtn,&tmp);
        }
    }
    free(tmp.pPcs);
}
\end{verbatim}}
The unary operation Short cut \textsc{Sc}:
{\small\begin{verbatim}
void scCalc(Pieces *rtn, Pieces *in){
    Searchy search;
    int x,oldRot,oldTrans,i;

    search.pPiece = in;
    search.number = -1;
    while((x = getNextTrue(&search))>= 0){
        oldRot      = getRot(x);
        oldTrans    = getTrans(x);

        if(oldRot < ROT_SLICES/2){
            for(i = ROT_SLICES/2; i <= ROT_SLICES/2+oldRot; i++){
                setA(rtn,i,oldTrans,1);
            }
        }else{
            for(i = oldRot-ROT_SLICES/2; i < ROT_SLICES/2; i++){
                setA(rtn,i,oldTrans,1);
}}}}
\end{verbatim}}
Test if two \fs{}s are equal.
{\small\begin{verbatim}
int testEqual(Pieces *pcs1, Pieces *pcs2){

    // set the unused bits at the end to 0    
    pcs1->pPcs[BYTES_USED-1] &= ~(255<<BIT_NUM(PIECES));
    pcs2->pPcs[BYTES_USED-1] &= ~(255<<BIT_NUM(PIECES));

    return (memcmp(pcs1->pPcs,pcs2->pPcs,BYTES_USED)==0);
}
\end{verbatim}}
A program that is successfully demonstrating the \textsc{Sc} operation, Pavlidis algorithm and the the composition.
{\small\begin{verbatim}
int main(int argc, char* argv[])
{
    // the FSPPs
    Pieces a,b,c,d,e,f;
    // testing if the defines are set correctly
    pretest();

    // allocate memory for the byte arrays
    initPieces(&a);
    initPieces(&b);
    initPieces(&c);
    initPieces(&d);
    initPieces(&e);
    initPieces(&f);
    
    // initialize an FSPP by spezifing atomic relations
    // using their orientation and distance index
    setA(&a,6,3,1);
    setA(&a,6,4,1);
    setA(&a,5,3,1);
    
    // initialize an FSPP using a metric coordinate system
    setPoint(&b,6,39,1);
    
    // initialize an FSPP by spezifing atomic relations
    // using their bit number in the array
    setB(&c,58,1);
    setB(&c,59,1);

    // ASCII output of the FSPPs
    printf(" FSPP a \n");
    printPieces(&a);
    printf(" FSPP b \n");
    printPieces(&b);
    printf(" FSPP c \n");
    printPieces(&c);

    // calculate the short cut of b - store the result in d
    unary(&d,&b,2);

    // print the Sc of b
    printf("\n\n result of SC b \n");
    printPieces(&d);

    // set all atomic relation to 0
    clearPiece(&d);
    
    // composition of a and b - store the result in d
    composition(&d,&a,&b);

    // print the result of the composition
    printf("\n\n composition result 1\n");
    printPieces(&d);

    // composition of a and d - store the result in e
    composition(&e,&a,&d);
    
    // do contour tracing and set the FSPP to the contour
    calcBoundaryPavlidi(&d);
    setPcsLikeBoundary(&d);

    // output of the contour
    printf("\n\n contour of composition \n");
    printPieces(&d);
    
    // composition with the contour
    composition(&f,&a,&d);

    // output to compare the compusition results with
    // and without contour
    printf("\n\n composition 2 (calculated without contour)\n");
    printPieces(&e);
    printf("\n\n composition 2 (calculated out contour)\n");
    printPieces(&f);
    
    // using a function to test for equality
    printf("\n e and f equal ? %d \n",testEqual(&e,&f));

    // do contour tracing and set the FSPP to the contour
    calcBoundaryPavlidi(&f);
    setPcsLikeBoundary(&f);

    // print the contour
    printf("\n\n contour of composition \n");
    printPieces(&f);
	
    // freeing the memory
    free(a.pPcs);
    free(b.pPcs);
    free(c.pPcs);
    free(d.pPcs);
    free(e.pPcs);
    free(f.pPcs);
}
\end{verbatim}}
This is the output of the program above (it uses 18 orientation distinctions and 20 distances).
\twocolumn 
\begin{center}

{\tiny\begin{verbatim}
 FSPP a
          TRANS
         00        10        20
         |    .    |    .    |
   000 : 00000000000000000000
 R 001 : 00000000000000000000
 O 002 : 00000000000000000000
 T 003 : 00000000000000000000
   004 : 00000000000000000000
   005 : 00010000000000000000
   006 : 00011000000000000000
   007 : 00000000000000000000
   008 : 00000000000000000000
   009 : 00000000000000000000
         |    .    |    .    |
   010 : 00000000000000000000
   011 : 00000000000000000000
   012 : 00000000000000000000
   013 : 00000000000000000000
   014 : 00000000000000000000
   015 : 00000000000000000000
   016 : 00000000000000000000
   017 : 00000000000000000000
 FSPP b
          TRANS
         00        10        20
         |    .    |    .    |
   000 : 00000000000000000000
 R 001 : 00000000000000000000
 O 002 : 00000000000000000000
 T 003 : 00000000000000000000
   004 : 00000000000000010000
   005 : 00000000000000000000
   006 : 00000000000000000000
   007 : 00000000000000000000
   008 : 00000000000000000000
   009 : 00000000000000000000
         |    .    |    .    |
   010 : 00000000000000000000
   011 : 00000000000000000000
   012 : 00000000000000000000
   013 : 00000000000000000000
   014 : 00000000000000000000
   015 : 00000000000000000000
   016 : 00000000000000000000
   017 : 00000000000000000000
 FSPP c
          TRANS
         00        10        20
         |    .    |    .    |
   000 : 00000000000000000000
 R 001 : 00000000000000000000
 O 002 : 00000000000000000000
 T 003 : 00000000000000000000
   004 : 00010000000000000000
   005 : 00010000000000000000
   006 : 00000000000000000000
   007 : 00000000000000000000
   008 : 00000000000000000000
   009 : 00000000000000000000
         |    .    |    .    |
   010 : 00000000000000000000
   011 : 00000000000000000000
   012 : 00000000000000000000
   013 : 00000000000000000000
   014 : 00000000000000000000
   015 : 00000000000000000000
   016 : 00000000000000000000
   017 : 00000000000000000000

 result of SC b
          TRANS
         00        10        20
         |    .    |    .    |
   000 : 00000000000000000000
 R 001 : 00000000000000000000
 O 002 : 00000000000000000000
 T 003 : 00000000000000000000
   004 : 00000000000000000000
   005 : 00000000000000000000
   006 : 00000000000000000000
   007 : 00000000000000000000
   008 : 00000000000000000000
   009 : 00000000000000010000
         |    .    |    .    |
   010 : 00000000000000010000
   011 : 00000000000000010000
   012 : 00000000000000010000
   013 : 00000000000000010000
   014 : 00000000000000000000
   015 : 00000000000000000000
   016 : 00000000000000000000
   017 : 00000000000000000000

 composition result 1
          TRANS
         00        10        20
         |    .    |    .    |
   000 : 00000000000000000000
 R 001 : 00000000000000000000
 O 002 : 00000000000000000000
 T 003 : 00000000000000000000
   004 : 00000000000000000000
   005 : 00000000000000000000
   006 : 00000000000000000000
   007 : 00000000000000000000
   008 : 00000000000000111100
   009 : 00000000000000111100
         |    .    |    .    |
   010 : 00000000000000111100
   011 : 00000000000000111100
   012 : 00000000000000000000
   013 : 00000000000000000000
   014 : 00000000000000000000
   015 : 00000000000000000000
   016 : 00000000000000000000
   017 : 00000000000000000000

 contour of composition
          TRANS
         00        10        20
         |    .    |    .    |
   000 : 00000000000000000000
 R 001 : 00000000000000000000
 O 002 : 00000000000000000000
 T 003 : 00000000000000000000
   004 : 00000000000000000000
   005 : 00000000000000000000
   006 : 00000000000000000000
   007 : 00000000000000000000
   008 : 00000000000000111100
   009 : 00000000000000100100
         |    .    |    .    |
   010 : 00000000000000100100
   011 : 00000000000000111100
   012 : 00000000000000000000
   013 : 00000000000000000000
   014 : 00000000000000000000
   015 : 00000000000000000000
   016 : 00000000000000000000
   017 : 00000000000000000000

 composition 2 (calculated without contour)
          TRANS
         00        10        20
         |    .    |    .    |
   000 : 00000000011111111000
 R 001 : 00000000001111111000
 O 002 : 00000000001111000000
 T 003 : 00000000000000000000
   004 : 00000000000000000000
   005 : 00000000000000000000
   006 : 00000000000000000000
   007 : 00000000000000000000
   008 : 00000000000000000000
   009 : 00000000000000000000
         |    .    |    .    |
   010 : 00000000000000000000
   011 : 00000000000000000000
   012 : 00000000001111110000
   013 : 00000000011111110000
   014 : 00000000011111110000
   015 : 00000000011111110000
   016 : 00000000011111111000
   017 : 00000000011111111000

 composition 2 (calculated out contour)
          TRANS
         00        10        20
         |    .    |    .    |
   000 : 00000000011111111000
 R 001 : 00000000001111111000
 O 002 : 00000000001111000000
 T 003 : 00000000000000000000
   004 : 00000000000000000000
   005 : 00000000000000000000
   006 : 00000000000000000000
   007 : 00000000000000000000
   008 : 00000000000000000000
   009 : 00000000000000000000
         |    .    |    .    |
   010 : 00000000000000000000
   011 : 00000000000000000000
   012 : 00000000001111110000
   013 : 00000000011111110000
   014 : 00000000011111110000
   015 : 00000000011111110000
   016 : 00000000011111111000
   017 : 00000000011111111000

 e and f equal ? 1

 contour of composition
          TRANS
         00        10        20
         |    .    |    .    |
   000 : 00000000010000001000
 R 001 : 00000000001000111000
 O 002 : 00000000001111000000
 T 003 : 00000000000000000000
   004 : 00000000000000000000
   005 : 00000000000000000000
   006 : 00000000000000000000
   007 : 00000000000000000000
   008 : 00000000000000000000
   009 : 00000000000000000000
         |    .    |    .    |
   010 : 00000000000000000000
   011 : 00000000000000000000
   012 : 00000000001111110000
   013 : 00000000010000010000
   014 : 00000000010000010000
   015 : 00000000010000010000
   016 : 00000000010000001000
   017 : 00000000010000001000\end{verbatim}}
\end{center}
\onecolumn


\section{Comparison}
\label{sec:comparison}
A comparison and classification of important calculi presented in this thesis is done in this section. This is done to help to choose the proper approach for spatial problems. It should also show, that the \FSPP~calculus developed in this thesis has unique properties which are useful in some context.  

\subsection{TPCC vs \fs}
TPCC, which is presented in Section \ref{sec:tpcc}, has a predetermined set of atomic relations. It differentiates two relative distances and eight orientations. Additionally four linear directions that are derived from the cardinal direction calculus (see Section \ref{subsubsection:orientation}) can be represented (straight front and back, left and right neutral) which give the calculus nice algebraic properties. For robotic applications these linear relations are less important because objects will almost certainly not be located directly on the straight line. The constant number of relatively few atomic relations enables a fast composition using table lookup. TCPP has already been used for robotic applications \cite{Dialog}. During the test series it became apparent, that, although it is often possible to achieve good results with TPCC, the calculus is too coarse for robust robotic applications. 

The advantages of \fs~over TPCC are quite straight forward - it supports more orientation distinctions and more distance distinctions than TPCC. In both calculi the relative orientation is defined over three points. \fs{}s absolute distance system is favorably for robotic representations while the relative distance in TPCC is better suited for interaction with humans. 

In applications such as described in \cite{Dialog} \fs~is expected to perform significantly better if a high granularity is chosen. Due to it's finer grain far less ambiguous situations are expected, that are situations in which more than one real life object is located in the acceptance areas of the relation. In situations where no real life object is found the \fs, the relation can be expanded by the conceptual neighbors of the atomic relations. This would not be possible with TPCC because it instantly would lead to very ambiguous results. 

\subsection{DOI vs \fs}

The \doi~presented in Section \ref{sec:doi} is a very mathematical calculus. The fact that is does not have atomic relations shows that it is not a qualitative approach. DOIs are successfully used in the robotic context, for example in maps using Voronoi graphs, as mentioned in Section \ref{sec:applications}. 

The \fs~uses the \doi~to calculate the composition. It is apparent that the composition algorithm of the \fs~has a bad complexity, especially compared to the DOI calculus who just needs one (DOI) calculation for it. But in this paper measures have been described to keep even high resolution \fs{}s usable for robotic applications. The main advantage of \fs~over the DOI approach is, that it is, to some extent, a qualitative calculus. Thus the fine properties described in Section \ref{sec:propertiesOfQSR} are still available for \fs. 
 
\subsection{\textsc{Gppc} vs \fs}
\textsc{Gppc} (see Section \ref{sec:gppc}) is quite similar to \fs. A minor difference is, that \textsc{Gppc} has fixed levels of granularity whereas the number of orientation and distance differentiations in \fs~can be chosen freely. \textsc{Gppc} thus limits the number of different resolution modes which might make it more user friendly. The granularity of \fs~on the other hand can easily be adjusted to the users needs, for example with many distances but few orientations.

The most important difference between the both is clearly, that \textsc{Gppc} uses relative distances while \fs~has an absolute distance system. The relative distance system of \textsc{Gppc} yields to quite accurate results of the unary operations while it causes more ambiguous composition relations. The situation for the \fs~is reverse. It has good composition results but the results of the unary operations are quite inexact. The composition is surely the more important operation, but I think that the disadvantages cancel each other out since both unary and binary operations are needed during constraint based reasoning. The relative distance system of \textsc{Gppc} is better suited for interactions with humans while the absolute distances used in \fs~have advantages for maps. 

\subsection{Classification of the calculi} 
\begin{figure}[ht]  
	\begin{captionabove}{Quantitative vs Qualitative calculi}
  	\label{fig:linie}
\begin{center}
		\includegraphics[height=1.9cm]{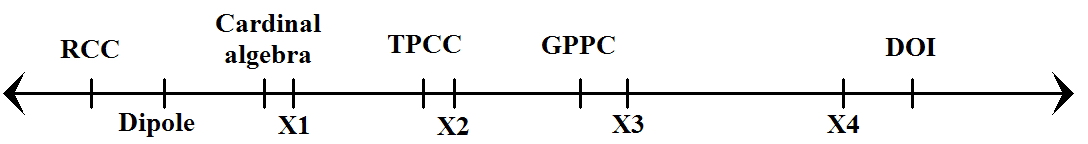}
\end{center}
	\end{captionabove}
\end{figure}

Only few calculi are absolutely qualitative. Many have more or less quantitative features. The transition from qualitative to quantitative representations is gradual. Figure \ref{fig:linie} shows a classification of different calculi on a line. More qualitative calculi are inserted on the left while the quantitative DOI calculus is on the right. The topological RCC is entirely qualitative, while TPCC can be seen as more number cumbering as the cardinal algebra because of the relatively higher number of basic relations. 

This classification also corresponds to the amount of positional information the calculi can represent (except of RCC which represents topological knowledge). Approaches to the left are more coarse than those on the right. \textsc{Gppc} and even more \fs~ don't have a fixed position on this line because of the different levels of granularity they offer. 

An extreme version of \fs~with just four orientations and one distance is more coarse than TPCC. It thus would have to be inserted at the X1 mark in the figure. A slightly finer \fs~ with, for example, eight orientations and three distances would be located at the X2 mark while even higher resolutions can be at the X3 mark. DOIs will always be more accurate than \fs~with even the highest level of granularity because they use real numbers for the orientation and the range. Therefore X4 indicates the rightmost mark possible for \fs. In this classification \textsc{Gppc} can be located in similar regions as the \fs.

%% file: chapters/conclusion.tex
%
%
%
%

\chapter{Conclusion}
\label{sec:conclusion}






\section{Summary}

In this diploma thesis a new calculus for reasoning about qualitative positional information with the name "\FSPP" (\fs) has been developed. The basic terms and definition have been defined in Section \ref{sec:rep1}. Representation and reasoning techniques as well as the properties of qualitative spatial reasoning are presented there. State of the art approaches have been outlined in Section \ref{sec:StateOfTheArt}. The \tpcc~(TPCC) is of special interest because not only the \fs~calculus is derived from it. The \doi~is important because it is needed for the reasoning process of \fs. The Granular Point Position Calculus is a recently developed approach that has many similarities with the one developed in this thesis.

The representation of the \fs~has been developed in Section \ref{sec:FSPP approach}. The absolute distance system and the relative orientation are outlined and merged into the ternary position representation. Unary and binary operations are needed for the reasoning process. The algorithms for both operations are presented. The composition makes use of the \doi~ while the unary inversion (\textsc{Inv}) operation makes use of the composition. The algorithm for the unary short cut (\textsc{Sc}) operation is presented so that the other operations can be calculated using \textsc{Inv} and \textsc{Sc}. The conceptual neighborhood concept can be applied to \fs, too.

An important improvement for the calculation of the composition is the use of a contour trancing algorithm beforehand. The terms that have their origin in computer graphics have been introduced and the application of the contour tracing algorithm of Theo Pavlidis to \fs~has been explained. The implementation of the \fs~calculus was shown to confirm the usability of this approach. The advantages and disadvantages of the \fs~compared to other important calculi have been pointed out and a classification was developed over them.

\section{Discussion and Outlook}

This thesis was written with the aim to develop a new calculus for qualitative spatial reasoning in the robotic context. An arbitrary level of granularity for representing positional knowledge was intended. I think that the \fs~approach at hand satisfies these expectations. It is defined over three points and offers a free number of distance and orientation distinctions. The relative orientation and absolute distance offer a good representation of positional knowledge. The necessary unary operations are defined. Because of the the absolute distance they are quite ambiguous. This is especially the case for those operations that are defined indirectly over the \textsc{Inv} and \textsc{Sc} operations. Future research could be done to define the unary \textsc{Sci}, \textsc{Hm} and \textsc{Hmi} operations directly to get better results during the reasoning process.

In \fs~ the composition is done using the DOI composition of the atomic relations. This leads to a huge number of DOI operations for very fine-grained representations. An effective way of reducing this number is presented by only using the border relations during the composition. During the extensive tests that have been done with the implementation of the \fs~it was shown that this border-composition is as exact as a composition with all basic relations. This is however not proved and thus a topic for future research. Related to this topic is the question, if 8-connectivity is really sufficient for the border-composition or if there are special cases in which 4-connectivity is required. 

Another interesting topic for future research is the question, which level of granularity and what distance systems for the \fs~representation are advisable for typical application scenarios. Especially the distance system is interesting and cognitive research and test series could be done to find representations that are optimal for interaction with humans as well as for usage in maps and other applications. 

The comparison with other state of the art calculi revealed that \fs~has unique properties which are useful in the robotic context. It also became apparent that the recently developed \textsc{Gppc} has many similarities with the \fs. It could be interesting to find algorithms that allow to transform knowledge represented with those calculi forth and back. That way the advantages of both approaches could be used.



%% file: appendix/TPCC.tex
%
%
%
%

\chapter{TPCC definition}
\label{cap:TPCCdef}
The formal definition of the TPCC relations are described by geometric configurations on the basis of a cartesian coordinate system represented by $R^2$.

Special Cases for $A = (x_A,y_A), B = (x_B,y_B), C = (x_c,y_c)$ :\\
$A, B \mbox{ dou } C := x_A = x_B \neq x_C \wedge y_A = y_B \neq y_C$\\
$A, B \mbox{ tri } C := x_A = x_B = x_C \wedge y_A = y_B = y_C$

For cases with $ A \neq B$ a relative radius $r_{A,B,C}$ and a relative angle $\phi_{A,B,C}$ are defined:

$r_{A,B,C} := \frac{\sqrt{\left(x_C-x_B\right)^2+\left(y_C-y_B\right)^2}}{\sqrt{\left(x_B-x_A\right)^2+\left(y_B-y_A\right)^2}}$

$\phi_{A,B,C} := \tan^{-1} \frac{y_C - y_B}{x_C - x_B} - \tan^{-1} \frac{y_B - y_A}{x_B - x_A}$

Spatial relations:

\begin{eqnarray*}
A, B \mbox{ sam } C & := & r_{A,B,C} = 0 \\
A, B \mbox{ csb } C & := & 0 < r_{A,B,C} < 1 \wedge \phi_{A,B,C}= 0 \\
A, B \mbox{ dsb } C & := & 1 \leq r_{A,B,C} \wedge \phi_{A,B,C}= 0 \\
A, B \mbox{ clb } C & := & 0 < r_{A,B,C} < 1 \wedge 0 < \phi_{A,B,C}\leq \pi/4 \\
A, B \mbox{ dlb } C & := & 1 \leq r_{A,B,C} \wedge 0 < \phi_{A,B,C}\leq \pi/4 \\
A, B \mbox{ cbl } C & := & 0 < r_{A,B,C} < 1 \wedge \pi/4 < \phi_{A,B,C} < \pi/2 \\
A, B \mbox{ dbl } C & := & 1 \leq r_{A,B,C} \wedge \pi/4 < \phi_{A,B,C} < \pi/2 \\
A, B \mbox{ csl } C & := & 0 < r_{A,B,C} < 1 \wedge \phi_{A,B,C} = \pi/2 \\
A, B \mbox{ dsl } C & := & 1 \leq r_{A,B,C} \wedge \phi_{A,B,C} = \pi/2 \\
A, B \mbox{ cfl } C & := & 0 < r_{A,B,C} < 1 \wedge \pi/2 < \phi_{A,B,C} < 3/4\pi \\
A, B \mbox{ dfl } C & := & 1 \leq r_{A,B,C} \wedge \pi/2 < \phi_{A,B,C} < 3/4\pi \\
A, B \mbox{ clf } C & := & 0 < r_{A,B,C} < 1 \wedge 3/4\pi \leq \phi_{A,B,C} < \pi \\
A, B \mbox{ dlf } C & := & 1 \leq r_{A,B,C} \wedge 3/4\pi \leq \phi_{A,B,C} < \pi \\
A, B \mbox{ csf } C & := & 0 < r_{A,B,C} < 1 \wedge \phi_{A,B,C}= \pi \\
A, B \mbox{ dsf } C & := & 1 \leq r_{A,B,C} \wedge \phi_{A,B,C}= \pi \\
A, B \mbox{ crf } C & := & 0 < r_{A,B,C} < 1 \wedge \pi < \phi_{A,B,C}\leq 5/4\pi \\
A, B \mbox{ drf } C & := & 1 \leq r_{A,B,C} \wedge \pi < \phi_{A,B,C}\leq 5/4\pi \\
A, B \mbox{ cfr } C & := & 0 < r_{A,B,C} < 1 \wedge 5/4\pi < \phi_{A,B,C} < 3/2\pi \\
A, B \mbox{ dfr } C & := & 1 \leq r_{A,B,C} \wedge 5/4\pi < \phi_{A,B,C} < 3/2\pi \\
A, B \mbox{ csr } C & := & 0 < r_{A,B,C} < 1 \wedge \phi_{A,B,C} = 3/2\pi \\
A, B \mbox{ dsr } C & := & 1 \leq r_{A,B,C} \wedge \phi_{A,B,C} = 3/2\pi \\
A, B \mbox{ cbr } C & := & 0 < r_{A,B,C} < 1 \wedge 3/2\pi < \phi_{A,B,C} < 7/4\pi \\
A, B \mbox{ dbr } C & := & 1 \leq r_{A,B,C} \wedge 3/2\pi < \phi_{A,B,C} < 7/4\pi \\
A, B \mbox{ crb } C & := & 0 < r_{A,B,C} < 1 \wedge 7/4\pi \leq \phi_{A,B,C} < 2\pi \\
A, B \mbox{ drb } C & := & 1 \leq r_{A,B,C} \wedge 7/4\pi \leq \phi_{A,B,C} < 2\pi \\
\end{eqnarray*}

%% file: appendix/DOI.tex
%
%
%
%

\chapter{DOI composition formula}
\label{cap:DOIcompform}

The resulting DOI $d_3$ of a composition of the DOIs $d_1$ and $d_2$ has its minimum and maximum radius and angle to be computed:

$min_{r_1^i,r_2^k,\phi_1^j,\phi_2^l} r_{\sum}\left(r_1^i,r_2^k,\phi_1^j,\phi_2^l\right)$\\
$max_{r_1^i,r_2^k,\phi_1^j,\phi_2^l} r_{\sum}\left(r_1^i,r_2^k,\phi_1^j,\phi_2^l\right)$\\
$min_{r_1^i,r_2^k,\phi_1^j,\phi_2^l} \phi_{\sum}\left(r_1^i,r_2^k,\phi_1^j,\phi_2^l\right)$\\
$max_{r_1^i,r_2^k,\phi_1^j,\phi_2^l} \phi_{\sum}\left(r_1^i,r_2^k,\phi_1^j,\phi_2^l\right)$

For $min_{r_1^i,r_2^k,\phi_1^j,\phi_2^l} r_{\sum}\left(r_1^i,r_2^k,\phi_1^j,\phi_2^l\right)$ 12 geometric cases have to be considered $r_3^1,\ldots,r_3^{12}$.
The $min_{r_1^i,r_2^k,\phi_1^j,\phi_2^l} r_{\sum}\left(r_1^i,r_2^k,\phi_1^j,\phi_2^l\right) = min\left(r_3^1,\ldots,r_3^{12}\right)$

\begin{eqnarray*}
r_3^1 & = & r_{\sum}\left(r_1^{min},r_2^{min},0,\phi_2^{min}\right)\\
r_3^2 & = & r_{\sum}\left(r_1^{min},r_2^{min},0,\phi_2^{max}\right)\\
r_3^3 & = & r_{\sum}\left(r_1^{min},r_2^{max},0,\phi_2^{min}\right)\\
r_3^4 & = & r_{\sum}\left(r_1^{min},r_2^{max},0,\phi_2^{max}\right)\\
r_3^5 & = & r_{\sum}\left(r_1^{max},r_2^{min},0,\phi_2^{min}\right)\\
r_3^6 & = & r_{\sum}\left(r_1^{max},r_2^{min},0,\phi_2^{max}\right)\\
r_3^7 & = & r_1^{min} - r_2^{max} \Leftarrow \left(\phi_2{min} \leq -\pi \leq \phi_2{max} \wedge r_1^{min} > r_2^{max}\right)\\
r_3^8 & = & r_2^{min} - r_1^{max} \Leftarrow \left(\phi_2{min} \leq -\pi \leq \phi_2{max} \wedge r_2^{min} > r_1^{max}\right)\\
r_3^9 & = & r_{\sum}\left(r_1^{min},r_1^{min}\cos\left(\pi-\phi_2^{max}\right),0,\phi_2^{max}\right) \Leftarrow \\ 
 &  & \left(\phi_2{max} > \frac{\pi}{2} \wedge r_2^{min} < r_1^{min}\cos\left(\pi-\phi_2^{max}\right)<r_2^{max}\right)\\
r_3^{10} & = & r_{\sum}\left(r_1^{min},r_1^{min}\cos\left(\pi+\phi_2^{min}\right),0,\phi_2^{min}\right) \Leftarrow\\
 & & \left(\phi_2{min} < -\frac{\pi}{2} \wedge r_2^{min} <  r_1^{min}\cos\left(\pi+\phi_2^{min}\right)<r_2^{max}\right)\\
r_3^{11} & = & r_{\sum}\left(-\cos\phi_2^{max}r_2^{min},r_2^{min},0,\phi_2^{max}\right) \Leftarrow \cos\phi_2{max}r_2^{min}+r_1^{min} < 0 < \cos\phi_2^{max}r_2^{min}+r_1^{max}\\
r_3^{12} & = & r_{\sum}\left(-\cos\phi_2^{min}r_2^{min},r_2^{min},0,\phi_2^{min}\right) \Leftarrow \cos\phi_2{min}r_2^{min}+r_1^{min} < 0 < \cos\phi_2^{min}r_2^{min}+r_1^{max}
\end{eqnarray*}

For $max_{r_1^i,r_2^k,\phi_1^j,\phi_2^l} r_{\sum}\left(r_1^i,r_2^k,\phi_1^j,\phi_2^l\right)$ the geometric analysis shows seven distinct cases over which the maximum $max\left(r_3^{13},\ldots,r_3^{19}\right)$ has to be computed:

\begin{eqnarray*}
r_3^{13} & = & r_{\sum}\left(r_1^{max},r_2^{min},0,\phi_2^{min}\right)\\
r_3^{14} & = & r_{\sum}\left(r_1^{max},r_2^{min},0,\phi_2^{max}\right)\\ 
r_3^{15} & = & r_{\sum}\left(r_1^{max},r_2^{max},0,\phi_2^{min}\right)\\ 
r_3^{16} & = & r_{\sum}\left(r_1^{max},r_2^{max},0,\phi_2^{max}\right)\\ 
r_3^{17} & = & r_{\sum}\left(r_1^{min},r_2^{max},0,\phi_2^{min}\right)\\ 
r_3^{18} & = & r_{\sum}\left(r_1^{min},r_2^{max},0,\phi_2^{max}\right)\\
r_3^{19} & = & r_{\sum}\left(r_1^{min},r_2^{max},0,0\right) \Leftarrow \phi_2^{min} < 0 < \phi_2^{max}\\ 
\end{eqnarray*}

The algorithm for $min_{r_1^i,r_2^k,\phi_1^j,\phi_2^l} \phi_{\sum}\left(r_1^i,r_2^k,\phi_1^j,\phi_2^l\right)$ and 
$max_{r_1^i,r_2^k,\phi_1^j,\phi_2^l} \phi_{\sum}\left(r_1^i,r_2^k,\phi_1^j,\phi_2^l\right)$ have been debugged. First the cases are calculated:

\begin{eqnarray*}
\phi_3^{1} & = & \phi_{\sum}\left(r_1^{min},r_2^{min},\phi_1^{min},\phi_2^{min}\right)\\
\phi_3^{2} & = & \phi_{\sum}\left(r_1^{min},r_2^{max},\phi_1^{min},\phi_2^{min}\right)\\
\phi_3^{3} & = & \phi_{\sum}\left(r_1^{max},r_2^{min},\phi_1^{min},\phi_2^{min}\right)\\
\phi_3^{4} & = & \phi_{\sum}\left(r_1^{max},r_2^{max},\phi_1^{min},\phi_2^{min}\right)\\
\phi_3^{5} & = & \phi_{\sum}\left(r_1^{min},r_2^{max},\phi_1^{min},\phi_2^{max}\right)\\
\phi_3^{6} & = & \phi_{\sum}\left(r_1^{min},r_2^{max},\phi_1^{min},-\frac{\pi}{2}-\sin^{-1}\frac{r_2^{max}}{r_1^{min}}\right)\Leftarrow \phi_2^{min} < -\frac{\pi}{2}-\sin^{-1}\frac{r_2^max}{r_1^{min}}<\phi_2^{max}\\
\phi_3^{7} & = & \phi_{\sum}\left(r_1^{min},r_2^{min},\phi_1^{max},\phi_2^{max}\right)\\
\phi_3^{8} & = & \phi_{\sum}\left(r_1^{min},r_2^{max},\phi_1^{max},\phi_2^{max}\right)\\
\phi_3^{9} & = & \phi_{\sum}\left(r_1^{max},r_2^{min},\phi_1^{max},\phi_2^{max}\right)\\
\phi_3^{10} & = & \phi_{\sum}\left(r_1^{max},r_2^{max},\phi_1^{max},\phi_2^{max}\right)\\
\phi_3^{11} & = & \phi_{\sum}\left(r_1^{min},r_2^{max},\phi_1^{max},\phi_2^{min}\right)\\
\phi_3^{12} & = & \phi_{\sum}\left(r_1^{min},r_2^{max},\phi_1^{max},\frac{\pi}{2}+\sin^{-1}\frac{r_2^{max}}{r_1^{min}}\right)\Leftarrow \phi_2^{min} < \frac{\pi}{2}+\sin^{-1}\frac{r_2^{max}}{r_1^{min}}<\phi_2^{max}\\
\phi_3^{13} & = & \phi_{\sum}\left(r_1^{max},r_2^{min},\phi_1^{max},\phi_2^{min}\right)\\
\phi_3^{14} & = & \phi_{\sum}\left(r_1^{max},r_2^{min},\phi_1^{min},\phi_2^{max}\right)\\
\end{eqnarray*}

The values of $\phi_3^{1-14}$ are modified to be in the range of $-2\pi \leq \phi_3^{n} \leq 0$.
After that $\phi_3^{1-14}$ are sorted by their value. The difference between the neighboring cases are calculated,  with the biggest and the smallest value being neighbors, too, since the values are angles on a circle! The values that have the biggest difference between each other are taken into account now - they are the new $\phi_3^{min}$ and $\phi_3^{max}$ - with the smaller value being min and the bigger being max. 

If the difference between the newly calculated $\phi_3^{min}$ and $\phi_3^{max}$ is greater than 180° the special case with $\phi^{max} = \pi$, $\phi^{min} = -\pi$ and $r^{min} = 0$ is set because the goal location can be at the reference point now! The same holds if the following conditions are met:
$\phi_2^{min} \leq -\pi \leq \phi_2^{max}$ or $r_1^{min} \leq r_2^{min}\leq r_1^{max}$ or $r_1^{min} \leq r_2^{min} \leq r_1^{max}$.

%% file: appendix/RCC.tex
\chapter{Topology definition}
\label{app:topology}

\textbf{Definition}: topology, topological space: Let $U$ be a non-empty set, the universe. A topology on $U$ is a family $T$ of subsets of $U$ that satisfies the following axioms:
\begin{enumerate}
    \item $U$ and $\emptyset$ belong to $T$,
    \item the union of any number of sets in $T$ belongs to $T$,
    \item the intersection of any two sets of $T$ belongs to $T$.
\end{enumerate}
A topological space is a pair $\left[U,T\right]$. The members of $T$ are called open sets.

In a topological space $\left[U,T\right]$, a subset $X$ of $U$ is called a closed set if its complement $X^c$ is an open set, i.e. if $X^c$ belongs to $T$. By applying the DeMorgan laws, we obtain the properties of closed sets:
\begin{enumerate}
    \item $U$ and $\emptyset$ are closed sets,
    \item the intersection of any number of closed sets is a closed set,
    \item the union of any two closed sets is a closed set.
\end{enumerate}
If the particular topology $T$ on a set $U$ is clear or not important, the $U$ can be referred to as the topological space.
Closely related to the concept of an open set is that of a neighborhood.

\textbf{Definition}: neighborhood, neighborhood system. Let $U$ be a topological space and $p \in U$ be a point in $U$.
\begin{itemize}
    \item $N \subset U$ is said to be a neighborhood of $p$ if there is an open subset $O \subset U$ such that $p \in O \subset N$.
    \item The family of all neighborhoods of $p$ is called the neighborhood system of $p$, denoted as $N_p$.
\end{itemize}
A neighborhood system $N_p$ has the property that every finite intersection of members of $N_p$ belongs to $N_p$. Based on the notion of neighborhood it is possible to define certain points and areas of a region.

Def interior, exterior, boundary, closure. Let $u$ be a topological space, $X \subset U$ be a subset of $U$ and $p \in U$ be a point in U.
\begin{itemize}
    \item $p$ is said to be an interior point of $X$ if there is a neighborhood $N$ of $p$ contained in $X$. The set of all interior points of $X$ is called the interior of $X$, denoted $i(X)$.
    \item $p$ is said to be an exterior point of $X$ of there is a neighborhood $N$ of $p$ that contains no point of $X$. The set of all exterior points of $X$ is called the exterior of $X$, denoted $e(X)$.
    \item $p$ is said to be a boundary point of $X$ if every neighborhood $N$ of $p$ contains at least one point in $X$ and one point is not in $X$. The set of all boundary points of $X$ is called the boundary of $X$, denoted $b(X)$.
    \item The closure of $X$, denoted $c(X)$, is the smallest closed set which contains $X$.
\end{itemize}

The closure of a set is equivalent to the union of its interior and its boundary. Every open set is equivalent to its interior, every closed set is equivalent to its closure.

Def regular open, regular closed. Let $x$ be a subset of a topological space $U$.
\begin{itemize}
    \item $X$ is said to be regular open if $X$ is equivalent to the interior of its closure, i.e. $X = i \left(c \left( X \right)\right)$.
    \item $X$ is said to be regular closed if $X$ is equivalent to the closure of its interior, i.e. $X = c \left(i \left( X \right)\right)$.
\end{itemize}

Two sets of a topological space are called separated if the closure of one set is disjoint from the other set, and vice-versa. A subset of a topological space is internally connected if it cannot be written as a union of two separated sets.

Topological spaces can be categorized according to how points or closed sets can be separated by open sets. Different possibilities are given by the separation axioms $T_i$. A topological space $U$ that satisfies axiom $T_i$ is called a $T_i$ space. Three of these separation axioms which are important for this work are the following:

$T_1$ : Given any two distinct points $p, q \in U$, each point belongs to an open set which does not contain the other point.
$T_2$ : Given any two distinct points $p, q \in U$, there exist disjoint open sets $O_p, O_q \subseteq U$ containing $p$ and $q$ respectively.
$T_3$ If $X$ is a closed subset of $U$ and $p$ is a point not in $X$, there exist disjoint open sets $O_X, O_p \subseteq U$ containing $X$ and $p$ respectively.

A connected space is a topological space which cannot be partitioned into two disjoint open sets, a topological space is regular, if it satisfies axioms $T_2$ and $T_3$.\par